%% file: example_paper.tex
\documentclass{article}
\usepackage{caption}

\usepackage{microtype}
\usepackage{graphicx, svg}
\usepackage{subfigure}
\usepackage{booktabs} 
\usepackage{xcolor, enumitem}
\usepackage{booktabs, multirow, makecell}
\usepackage{amsthm}
\usepackage[T1]{fontenc}
\usepackage{hyperref}
\usepackage{hyperref}
\usepackage{url}
\usepackage{graphicx}

\usepackage[accepted]{icml2026}

\makeatletter
\renewcommand{\Notice@String}{Preprint.} 
\makeatother

\usepackage{amsmath}
\usepackage{amssymb}
\usepackage{mathtools}
\usepackage{amsthm}

\usepackage[capitalize,noabbrev]{cleveref}

\theoremstyle{plain}

\theoremstyle{definition}

\theoremstyle{remark}

\usepackage[textsize=tiny]{todonotes}

\icmltitlerunning{Preprint}

\begin{document}

\twocolumn[
\icmltitle{
LASS-ODE: Scaling ODE Computations to Connect Foundation Models with Dynamical Physical Systems
}



\icmlsetsymbol{equal}{*}

\begin{icmlauthorlist}
\icmlauthor{Haoran Li}{yyy}
\icmlauthor{Chenhan Xiao}{yyy}
\icmlauthor{Lihao Mai}{yyy}
\icmlauthor{Yang Weng}{yyy}
\icmlauthor{Erik Blasch}{comp}
\end{icmlauthorlist}

\icmlaffiliation{yyy}{Department of Electrical, Computer, and Energy Engineering, Arizona State University, AZ, United States}
\icmlaffiliation{comp}{Air Force Research Laboratory, VA, United States}

\icmlcorrespondingauthor{Haoran Li}{lhaoran@asu.edu}


\vskip 0.3in
]



\printAffiliationsAndNotice{}  


\begin{abstract}
Foundation models have transformed language, vision, and time series data analysis, yet progress on dynamic predictions for physical systems remains limited. Given the complexity of physical constraints, two challenges stand out. $(i)$ Physics-computation scalability: physics-informed learning can enforce physical regularization, but its computation (e.g., ODE integration) does not scale to extensive systems. $(ii)$ Knowledge-sharing efficiency: the attention mechanism is primarily computed within each system, which limits the extraction of shared ODE structures across systems. We show that enforcing ODE consistency does not require expensive nonlinear integration: a token-wise locally linear ODE representation preserves physical fidelity while scaling to foundation-model regimes. Thus, we propose novel token representations that respect locally linear ODE evolution. Such linearity substantially accelerates integration while accurately approximating the local data manifold. Second, we introduce a simple yet effective inter-system attention that augments attention with a common structure hub (CSH) that stores shared tokens and aggregates knowledge across systems. The resulting model, termed LASS-ODE (\underline{LA}rge-\underline{S}cale \underline{S}mall \underline{ODE}), is pretrained on our $40$GB ODE trajectory collections to enable strong in-domain performance, zero-shot generalization across diverse ODE systems, and additional improvements through fine-tuning. Our code is available at \href{https://anonymous.4open.science/r/LASS-ODE-B25A}{LASS-ODE}.
\end{abstract}

\section{Introduction}
Foundation models have achieved transformative progress across modalities such as text, vision, time series, and graphs \citep{vaswani2017attention,devlin2019bert,brown2020language,dosovitskiy2020image,kaplan2020scaling,das2024decoder,xiaoming2025time,liu2025sundial,mao2024position,liu2025graph}. At the core lies an attention mechanism, which enables efficient global and dynamic relation modeling. Empirically, transformer-based foundation models exhibit predictable scaling behavior: performance increases smoothly along three dimensions: data volume, model size, and compute \citep{kaplan2020scaling}. This scaling law lays a solid foundation for improving model capability through the increased scale and motivates the pursuit of large foundation models for physical systems.

However, existing foundation models remain difficult to deploy for dynamic predictions in many scientific and engineering domains whose behavior is governed by ordinary differential equations (ODEs). Such dynamics underpin core functions including simulation, forecasting, digital twins, and control across engineered systems \citep{vallado2001fundamentals,rasheed2019digital,10636963,10891664}, environmental processes \citep{lorenz2017deterministic}, and health systems \citep{qian2021integrating}. For example, while time-series foundation models can ingest sequential numerical observations \citep{nie2022time,das2024decoder,ansari2024chronos}, they primarily learn shared temporal regularities such as seasonality and cycles, rather than capturing the governing ODEs.


Small-scale approaches can directly model ODEs within a single system; their common approaches are symbolic regression, Physics-informed neural networks (PINNs), and Neural ODEs. Symbolic regression infers explicit governing equations from data, producing closed-form derivatives that can be integrated at arbitrary temporal resolutions \citep{brunton2016discovering,li2022console}. PINNs embed physical constraints into coordinate-based neural fields \citep{cuomo2022scientific}. The Neural ODE method instead learns the underlying vector field using a neural network \citep{chen2018node,rubanova2019latent,kidger2020neural,li2025neural}. These approaches can achieve high accuracy on targeted systems, yet often require case-specific tuning. More importantly, they do not generalize across different systems, even though many systems share ODE structures such as energy-conserving Hamiltonian forms \citep{greydanus2019hamiltonian}, damping and forcing patterns, or oscillatory dynamics \citep{krishnapriyan2021characterizing,rackauckas2020universal}.

Inspired by these advances, can we directly integrate diverse ODE constraints into a single foundation model? In general, existing paradigms fall short. Enforcing ODE consistency incurs substantial physics-side computation, and this overhead does not scale to the rich training systems. Moreover, effective learning requires extracting shared dynamical structure across systems to enable efficient transfer and generalization. Many pioneering LLMs do not emphasize ODE-constraint design because their primary tasks are typically intra-context. While some recent work explores cross-attention for aggregating information across multiple inputs \citep{jaegle2021perceiver,lopes2023cross} and retrieval-based augmentation \citep{borgeaud2022improving}, they often introduce additional architectural machinery and/or nontrivial compute and memory overhead (see Section \ref{sec:related}).

To address this gap, we propose the fourth scaling: physics-computation scaling through a new philosophy, \emph{LASS (LArge-Scale Small) ODE}. LASS-ODE first \emph{scales down} to construct physics-aware, computation-scalable tokens, and then \emph{scales up} through intra- and inter-system attention to identify shared knowledge and enhance generalization during pretraining. Specifically, in the first phase, LASS-ODE linearizes the ODE evolution within each token by learning a local tangent-space approximation of the ODE data (i.e., state) manifold. Composing these piecewise-linear tokens yields an accurate approximation of the overall nonlinear dynamics \citep{khalil2002nonlinear}. In the second phase, we design an extremely simple yet effective intra- and inter-system attention mechanism. We introduce a parameterized common structure hub (CSH), a set of global tokens that memorizes shared dynamical structure across systems. During training, LASS-ODE concatenates the CSH tokens with each system’s tokens and apply standard self-attention over the augmented sequence. This design avoids specialized modules and extensive hyperparameter tuning for balancing intra- versus inter-system interactions, and incurs only a small, fixed overhead compared with retrieval pipelines, while delivering strong empirical performance. In general, our contributions are threefold:
\begin{itemize}[itemsep=2pt, topsep=4pt]
\item \textbf{Scalable ODE-Constrained Modeling with Inter-System Attention.} We introduce token-wise linear ODE as a scalable alternative to Neural ODE (Section \ref{subsec:linear_ode_estimate}) and develop inter-system attention for common structure sharing across ODEs (Section \ref{subsec:intra-inter-attention}).

\item \textbf{Validated Building Blocks.} We complement the core method with a set of components that are necessary for stable training and accurate prediction, including channel-independent processing, radial basis function (RBF) time embedding and modulation, channel encoding, and Mixture of Experts (MoEs). We quantify their effects via ablations (Sections~\ref{subsec:ode_tokenization}--\ref{subsec:intra-inter-attention} and \ref{subsec:ablation}).

\item \textbf{Heterogeneous Data Handling and Large Empirical Validation.} We curate a diverse, large-scale dataset of ODE trajectories for pretraining, spanning heterogeneous state dimensions, sampling resolutions, and time scales. We preprocess all trajectories into a normalized space to enable stable and scalable training (Section \ref{subsec:data_handle}). Experiments demonstrate strong extrapolation performance and zero-shot generalization across diverse ODE systems. For challenging systems, simple Low-Rank Adaptation (LoRA) \citep{hu2022lora} can further yield significant improvements.
\end{itemize}

\section{Preliminaries}
\label{sec:preliminaries}
We consider families of dynamical systems governed by ODEs.  
For a given system with the state $\boldsymbol{x} \in \mathcal{X}$, where $\mathcal{X}$ denotes the system's state space, its evolution is described by a vector field \( f: \mathcal{X} \rightarrow \mathcal{X} \) as $\dot{\boldsymbol{x}} = f(\boldsymbol{x})$. We define the problem as follows. 

\begin{itemize}[itemsep=2pt, topsep=4pt]
\item \textbf{Given}: Sequential state observations $\{\boldsymbol{x}_i(t^i_j)\}_{i,j}$ from multiple systems indexed by $i$, each with distinct state spaces $\mathcal{X}_i$. For the $i^{th}$ system, the timestamps of all observed points \( \mathcal{T}^i_{\text{obs}} := \{ t^i_j \}_j \) are also available. 
\item \textbf{Dataset Heterogeneity}: The given ODE systems may vary in both state dimension and temporal resolution, with timescales ranging from milliseconds to years. We show how to handle them in Sections \ref{subsec:data_handle} and \ref{subsec:ode_tokenization}.
\item \textbf{Find}: A foundation model that is consistent with the underlying ODE $f$ and can accurately extrapolate full trajectories by conditioning on a short observed prefix.
\end{itemize}

\begin{figure*}[t]
    \centering
  \includegraphics[width=1\linewidth]{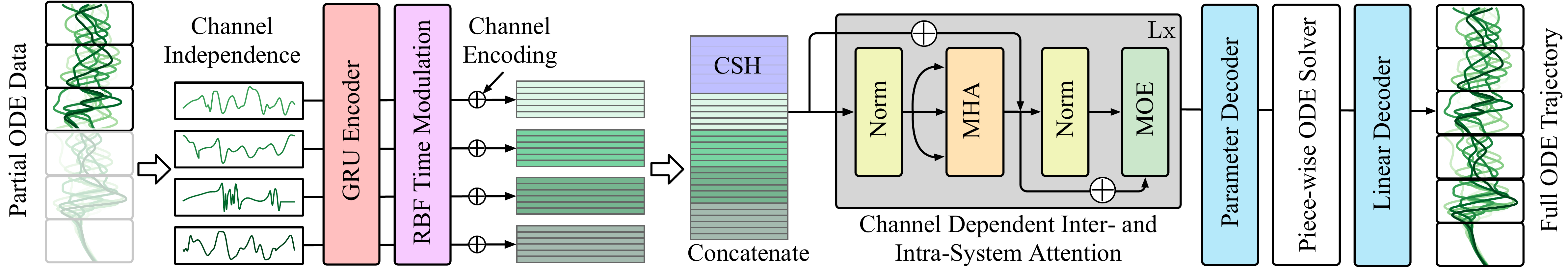}
    \caption{
    The proposed LASS-ODE framework.
    }
    \label{fig:bigpic}
    
\end{figure*}

    
In the following, we introduce some preliminaries and omit the system index $i$ for notational simplicity.

\textbf{Latent ODE to Enforce ODE Constraints.} Latent ODEs \citep{rubanova2019latent} and Neural ODEs \citep{chen2018node} naturally achieve ODE-consistent trajectory decoding through differentiable integration of a learned vector field. Moreover, their computation can be amenable to large-scale training by executing batched ODE integration on a GPU, leveraging CUDA tensor operations for efficient throughput. Specifically, in Latent ODE, an encoder $f_{\text{enc}}$ processes $N$ observations $\{\boldsymbol{x}(t_i)\}_{i =0}^N$ and \( \mathcal{T}_{\text{obs}} := \{ t_i \}_{i=0}^N \) to infer an initial latent state $\boldsymbol{z}(t_0)\in \mathcal{Z}$ that summarizes past information. The posterior distribution over this state is parameterized as: $\boldsymbol{z}(t_0) \sim \mathcal{N}\big(f_{\text{enc}}(\{\boldsymbol{x}(t_i)\}, \mathcal{T}_{\text{obs}})\big)$. Given $\boldsymbol{z}(t_0)$, the decoder evolves the latent trajectory whose derivative is  parameterized by a neural network $h_\theta$, and predicts the observed states using a likelihood model $p_\theta$: $\boldsymbol{z}(t_i) = \mathrm{ODESolve}(h_\theta, \boldsymbol{z}(t_0), t_0, t_i),\ \ \boldsymbol{x}(t_i) \sim p_\theta(\boldsymbol{x}(t_i)\mid \boldsymbol{z}(t_i))$. 

Here, $\mathrm{ODESolve}(h_\theta, \boldsymbol{z}(t_0), t_0, t_i)$ solves the initial value problem (IVP) defined by the neural ODE 
$\dot{\boldsymbol{z}}(t) = h_\theta(\boldsymbol{z}(t))$ with initial condition $\boldsymbol{z}(t_0)$, and evaluates the latent state at time $t_i$. $p_\theta$ as a multi-layer perceptron (MLP) decodes the latent state back to the observed state. Training maximizes the variational lower bound (ELBO) with KL (Kullback-Leibler) divergence: $\text{ELBO} = \mathbb{E}_{\boldsymbol{z}(t_0)} \left[ \log p_\theta(\{\boldsymbol{x}(t_i)\}) \right] -\text{KL}\left[\mathcal{N}\big(f_{\text{enc}}(\{\boldsymbol{x}(t_i)\}, \mathcal{T}_{\text{obs}})\big) \,\|\, p(\boldsymbol{z}(t_0))\right],$ where $p(\boldsymbol{z}(t_0))$ is typically a standard Gaussian prior.

\textbf{Multi-Head Attention (MHA)}. Given query \( Q \), key \( K \), and value \( V \), we have: 
\begin{equation}
\begin{aligned}
\label{eqn:MHA}
\small
&\mathrm{MHA}(Q, K, V) = \mathrm{Concat}(\text{head}_1, \dots, \text{head}_h) W^O, \\
&\text{head}_i = \mathrm{Softmax}( Q W_i^Q (K W_i^K)^\top/\sqrt{d_k} ) V W_i^V,
\end{aligned}
\end{equation}
where \( d_k \) is the dimension of each query/key vector, $W^O$, $W_i^Q$, and $W_i^K$ are weight matrices.

\section{LASS-ODE: Large-Scale Small ODE}

Fig.~\ref{fig:bigpic} illustrates the overall LASS-ODE framework, which comprises three components: (1) ODE trajectory tokenization, including a GRU encoder, RBF-based time embedding with modulation, and channel encoding; (2) a Transformer backbone with intra- and inter-system attention and MoE modules; and (3) a scalable ODE-constrained decoder that evolves latent features via piecewise-linear ODE integration. 

\subsection{Unified Data Handling}
\label{subsec:data_handle}

\textbf{Channel Independence for Multi-Dimensional Systems}. We adopt channel-independent processing \citep{nie2022time}. Specifically, each channel (i.e., the $i^{th}$ component of $\boldsymbol{x}$, denoted by $\boldsymbol{x}^{(i)}$) is processed independently during data pre-processing and tokenization. This may appear counterintuitive since the components of $\boldsymbol{x}$ are intrinsically coupled through the underlying ODEs. We nevertheless find this design beneficial for two reasons: (i) it better accommodates heterogeneous components (e.g., differing patterns) by avoiding premature entanglement, while cross-channel correlations are subsequently captured by the attention module; and (ii) it supports ODE systems with variable state dimensions, as the model processes channels one by one.

\textbf{Value/Time Normalization and Multi-timescale Control}. For stable training across datasets, we normalize each channel to the range $[-1,1]$ and rescale the time axis to $[0,1]$. We obtain normalized time by dividing the original timestamps by a system-specific maximum time $t_{\text{max}}$. For systems operating at different time scales, we choose different maximum times so that trajectories have a comparable effective frequency (i.e., similar numbers of oscillation cycles) within the unit interval. For example, we use $t_{\text{max}}=30$s for second-level data, while using $t_{\text{max}}=3$s for sub-second data. Intuitively, this zooms into sub-second dynamics by using a shorter window as the unit time, reducing frequency disparities across systems and stabilizing training. A trade-off is that for faster dynamics, the physical rollout window ($[t_0,t_{\text{max}}]$) becomes shorter. Longer extrapolation can be obtained by autoregressive rollout over multiple windows.

\textbf{Remark}: In our extrapolation setting, the target horizon typically extends beyond the last observed time $t_N$ (i.e., $t_{\text{max}}>t_N$). Accordingly, we condition on a prefix of the observations and predict the remaining trajectory over the full normalized interval $[0,1]$. The left and the right parts of Fig. \ref{fig:bigpic} give an example of the input and output, respectively.

\subsection{Time-Aware Tokenization for Multi-Resolution}
\label{subsec:ode_tokenization}

To prepare inputs for the Transformer backbone, we tokenize each ODE trajectory into a sequence of token embeddings that should ultimately support preserving the ODE structure. In Section~\ref{subsec:linear_ode_estimate}, we use the token embeddings (after attention) to estimate the token-wise ODE parameters, and then solve the resulting IVP. This requires the embeddings to be \emph{temporally coherent} so that the inferred parameters are consistent across tokens and yield a stable rollout. In contrast, prior work \citep{nie2022time,liu2024timer,liu2025sundial} typically applies an MLP to each local segment of raw signals independently, producing weakly coupled embeddings that can lead to inconsistent token-wise ODE parameter estimates. 
To mitigate this inconsistency, we condition all token embeddings on a shared trajectory-level temporal summary generated by a sequence encoder (e.g., a GRU).

\textbf{Time-Aware gated recurrent unit (GRU) Encoding}. As different ODE datasets may exhibit multiple sampling resolutions, we adopt a time-aware GRU encoder (the light pink box in the left of Fig. \ref{fig:bigpic}). For the $j^{th}$ channel, the encoder processes the concatenated input $[\boldsymbol{x}^{(j)}(t), \Delta t]$, where $\Delta t$ denotes the sampling interval (after normalization) of the specific ODE system. Specifically, we have:
\begin{equation}
\label{eq:gru_encoder} 
\boldsymbol{h}^{(j)} = \mathrm{GRU}(\{[\boldsymbol{x}^{(j)}(t_i);\Delta t_i]\}_{i=0}^N).
\end{equation}
Other options for encoding multi-resolution data, such as RNN-$\Delta_t$ \citep{che2018recurrent} and ODE-RNN \citep{rubanova2019latent}, incur higher computational costs. 

Given the global temporal feature $\boldsymbol{h}^{(j)}$, we aim to form a token embedding for each raw data segment. Recall that the GRU embedding is used to generate token-wise ODE parameters; therefore, it must be anchored to the segment's timestamp and focus on the local and neighboring temporal information. However, standard time (or position) encodings have two limitations in our setting: $(1)$ methods such as sinusoidal encodings \citep{vaswani2017attention} and rotary position embeddings (RoPE) \citep{su2024roformer} mainly represent relative phase but do not explicitly enforce locality or distance-based decay to emphasize local and neighboring segments, and $(2)$ they typically incorporate time by additive fusion with the token embedding, which is less effective when we only have $\boldsymbol{h}^{(j)}$.

\textbf{RBF Time Modulation-based Tokenization}. We propose to employ an RBF-based time encoding and a time-conditioned modulation scheme (instead of additive fusion) to cope with $\boldsymbol{h}^{(j)}$ and produce local embeddings. RBF kernels provide \emph{localized} activations in time: a time instant is encoded by a bank of kernels with different centers, where each kernel measures how close the instant is to its center with smooth distance-based decay. Such locality-aware encoding is crucial for capturing local ODE parameters.

Specifically, we partition the normalized trajectory on $[0,1]$ into $K_{\text{token}}$ non-overlapping tokens, where the $k^{th}$ token corresponds to the temporal segment at $[t_k^{\text{start}},\,t_k^{\text{end}}]$. To encode where the $k^{th}$ token sits along the trajectory, we use an RBF time embedding evaluated at its start time $t_k^{\text{start}}$: we place $K_{\text{rbf}}$ Gaussian kernels with centers $\{\mu_\ell\}_{\ell=1}^{K_{\text{rbf}}}$ evenly spaced on $[0,1]$, and stack their responses into a vector (with bandwidth $\sigma$). The $\ell^{th}$ kernel response is: $\phi_{\text{rbf}}^{(\ell)}\!\left(t_k^{\text{start}}\right)
=\exp(-\frac{\left(t_k^{\text{start}}-\mu_\ell\right)^2}{2\sigma^2})$. Instead of adding $\phi_{\text{rbf}}(t_k^{\text{start}})$ to $\boldsymbol{h}^{(j)}$ to form the embedding for the $j^{th}$ channel and $k^{th}$ token, we use $\phi_{\text{rbf}}(t_k^{\text{start}})$ to \emph{modulate} $\boldsymbol{h}^{(j)}$. Basically, modulation \citep{perez2018film} provides a stronger time-conditioned scaling and shifting to induce token-specific locality. Specifically, we have:
\begin{equation}
\begin{aligned}
\label{eq:modulation}
\big(\boldsymbol{\gamma}_k;\boldsymbol{\beta}_k\big)
&=\mathrm{MLP}\!\Big(\mathrm{LN}\big(\phi_{\text{rbf}}(t_k^{\text{start}})\big)\Big),\\
\boldsymbol{e}_{jk}
&=\boldsymbol{\gamma}_k \odot \mathrm{MLP}\!\big(\boldsymbol{h}^{(j)}\big)+\boldsymbol{\beta}_k,
\end{aligned}
\end{equation}
where $\boldsymbol{\gamma}_k,\boldsymbol{\beta}_k,\boldsymbol{e}_{jk}\in\mathbb{R}^{d_{\text{model}}}$, and $d_{\text{model}}$ is the dimension of the embedding. MLPs are used to align dimensions, while the layer norm (LN) standardizes RBF features and makes training more stable. $\odot$ is the element-wise multiplication.

\textbf{Channel Encoding}. 
To further locate different channels in an ODE system, we use a learnable lookup table $C\in\mathbb{R}^{d^{\text{max}}_x\times d_{\text{model}}}$, where the $j^{th}$ row $C^{(j)}\in\mathbb{R}^{d_{\text{model}}}$ provides the channel encoding for channel $j$. $d^{\text{max}}_x$ is the maximum number of state channels across all ODE systems. We then add this vector to $\boldsymbol{e}_{jk}$ to obtain the final embedding for channel $j$ and token $k$: $\tilde{\boldsymbol{e}}_{jk} \;=\; \boldsymbol{e}_{jk} + C^{(j)}$. Although we only have a prefix (with $t_N\leq t_{\text{max}}$), our tokenization still produces $\tilde{\boldsymbol{e}}_{jk}$ for all $K_{\text{token}}$ tokens over $[t_0,t_{\text{max}}]$ (i.e., the normalized interval $[0,1]$). This is because the time modulation in Eq.~\eqref{eq:modulation} depends only on the token time $t_k^{\text{start}}$ and $\boldsymbol{h}^{(j)}$, enabling the model to instantiate embeddings for unobserved segments. Section \ref{subsec:ablation} verifies these components.

\subsection{Intra- and Inter-System Attention for Efficiency}
\label{subsec:intra-inter-attention}

\textbf{Intra-System Attention: Self-Attention without A Causal Mask}. We stack all channel--token embeddings $\tilde{\boldsymbol{e}}_{jk}\in\mathbb{R}^{d_{\text{model}}}$ (for $j=1,\dots,d_x$ and $k=1,\dots,K_{\text{token}}$) into a single embedding matrix $\tilde{E}\in\mathbb{R}^{(K_{\text{token}}d_x)\times d_{\text{model}}}$, where $d_x$ is the dimension of the ODE state $\boldsymbol{x}(t)$. We then apply self-attention to $\tilde{E}$ without a causal mask. This drives tokens to attend \emph{across channels} to capture ODE state coupling and \emph{across all tokens} on $[0,1]$, because the instantaneous state can be insufficient to determine the current dynamics under partial observability or delayed coupling (e.g., $\dot{\boldsymbol{x}}(t)$ may depend on $\boldsymbol{x}(t-\Delta)$ \citep{hale2013introduction}). 


\textbf{Integrated Attention with a Common Structure Hub (CSH)}. 
We introduce a learnable matrix $E_{\text{CSH}}\in \mathbb{R}^{K_{\text{CSH}}\times d_{\text{model}}}$ that stores $K_{\text{CSH}}$ shared tokens encoding common dynamical structures across ODE systems (e.g., oscillations, damping, and saturation). During training, each system’s tokens attend to these global tokens, enabling the model to retrieve reusable dynamical evidence. As illustrated by the middle gray box in Fig.~\ref{fig:bigpic}, LASS-ODE treats the training tokens and the CSH tokens equally and applies the following self-attention:
\begin{equation}
\begin{aligned}
\label{eq:inter_intra_attention}
E_0 &= (\tilde{E};\,E_{\text{CSH}})\in\mathbb{R}^{(K_{\text{token}}d_x+K_{\text{CSH}})\times d_{\text{model}}},\\
E_1 &= E_0+ \mathrm{MHA}\big(\mathrm{LN}(E_0),\,\mathrm{LN}(E_0),\,\mathrm{LN}(E_0)\big),\\
E_2 &= E_1 + \mathrm{MoE}\big(\mathrm{LN}(E_1)\big),\\
\tilde{E}_{\text{next}} &=E_2^{(1:K_{\text{token}}d_x)}\in\mathbb{R}^{(K_{\text{token}}d_x)\times d_{\text{model}}},
\end{aligned}
\end{equation} 
where $\tilde{E}_{\text{next}}$ takes the first $K_{\mathrm{token}}d_x$ rows as the next-layer input features. We adopt MoE to let different experts specialize in dynamical patterns (e.g., oscillatory vs. damping). 
We denote by $\tilde{E}_{\text{final}}\in\mathbb{R}^{(K_{\text{token}}d_x)\times d_{\text{model}}}$ the output token features after stacking $L$ layers of integrated self-attention (see the middle of Fig.~\ref{fig:bigpic}).

\subsection{Token-Wise Linear ODE Decoding for Scalability} 
\label{subsec:linear_ode_estimate}

In this subsection, we describe how to use $\tilde{E}_{\text{final}}$ to reconstruct an ODE-constrained trajectory, and more importantly, how to do so in a scalable manner. LASS-ODE follows the Latent ODE paradigm and solves the IVP (see Section \ref{sec:preliminaries}) to obtain a latent flow, which is then decoded to the state space. Specifically, the initial condition in the IVP should be characterized by the whole observed data. Since Eq. \eqref{eq:gru_encoder} computes $\boldsymbol{h}^{(j)}$, we have:
\(
\boldsymbol{z}^{(j)}_0 \sim \mathcal{N}\big(g_{\mu}(\boldsymbol{h}^{(j)}),g_{\sigma}(\boldsymbol{h}^{(j)})\big).
\)
where $g_{\mu}$ and $g_{\sigma}$ are MLPs to convert $\boldsymbol{h}^{(j)}$ into the mean and the variance of the initial latent state $\boldsymbol{z}^{(j)}_0$.

Latent ODE typically solves the IVP with a nonlinear MLP derivative, which can be computationally expensive (see Table \ref{tab:cost_compare}). Since LASS-ODE already obtains high-level tokenized representations, it is natural to assume a locally linear ODE flow within each token interval. Specifically, we define a channel- and token-specific linear dynamics
$\dot{\boldsymbol{z}}^{(j)}(t)=A_{jk}\boldsymbol{z}^{(j)}(t)+\boldsymbol{b}_{jk}$ for $t\in[t_k^{\text{start}},t_k^{\text{end}}]$, where $A_{jk}\in\mathbb{R}^{d_{z}\times d_{z}}$ and $\boldsymbol{b}_{jk}\in\mathbb{R}^{d_{z}}$ are ODE parameters, and $d_z$ is the latent space dimensionality. These parameters can be obtained by a parameter decoder (an MLP) \( f_{\text{param}} \), as shown in the blue box in the right part of Fig. \ref{fig:bigpic}. Then, we have:
\begin{equation}
\label{eq:param_estimator}
    (A_{jk}, \boldsymbol{b}_{jk}) = f_{\text{param}}(\tilde{E}_{\text{final}}^{(jk)}),
\end{equation}
where $\tilde{E}_{\text{final}}^{(jk)}\in\mathbb{R}^{d_{\text{model}}}$ is the final feature vector for the $j^{th}$ channel and the $k^{th}$ token. Essentially, Eq. \eqref{eq:param_estimator} estimates the token-wise tangent space of the ODE manifold. As discussed in \citep{khalil2002nonlinear}, under standard smoothness conditions, such local linearization can approximate nonlinear ODE dynamics over a sufficiently small neighborhood. Finally, the IVP is solved as follows:
\begin{align}
&h^{(j)}(\boldsymbol{z}^{(j)}(t)) := A_{jk}(t)\boldsymbol{z}^{(j)}(t) + \boldsymbol{b}_{jk}(t), \nonumber \\
&\text{where } A_{jk}(t) := A_{jk}, \ \boldsymbol{b}_{jk}(t) := \boldsymbol{b}_{jk} \quad \text{if } t \in [t_k^{\text{start}}, t_k^{\text{end}}], \nonumber \\
&\boldsymbol{z}^{(j)}(t) = \mathrm{ODESolve}\left( h^{(j)}(\boldsymbol{z}^{(j)}(t)), \ \boldsymbol{z}^{(j)}_0, t\in[0,1]\right), \nonumber \\
&\hat{\boldsymbol{x}}^{(j)}(t) = W_{\text{dec}}\boldsymbol{z}^{(j)}(t) + b_{\text{dec}}, \label{eq:piecewise_odesolve}
\end{align}
where we define the token-wise linear function $h^{(j)}(\boldsymbol{z}^{(j)}(t))$ as the derivative. The white box in the right part of Fig. \ref{fig:bigpic} is the introduced solver. After obtaining the flow $\boldsymbol{z}^{(j)}(t)$, a linear decoder with weight $W_{\text{dec}}$ and bias $d_{\text{dec}}$ (see the blue box on the right of Fig.~\ref{fig:bigpic}) reconstructs the full trajectory $\hat{\boldsymbol{x}}^{(j)}(t)$, which by construction lies on an ODE manifold. 

\textbf{ODE Integration Cost Comparison}. For fair comparison, we consider a Latent ODE baseline whose latent derivative is parameterized by a nonlinear MLP, i.e., $\dot{\boldsymbol{z}} = h(\boldsymbol{z};\tilde{E}_{\text{final}}^{(jk)})$. We compare it with our piecewise-linear ODE. We assume both $h(\cdot)$ and $f_{\text{param}}(\cdot)$ (in Eq. \eqref{eq:param_estimator}) are $L$-layer MLPs with the same hidden width $K_{\text{width}}$. Let $n_{\text{step}}$ denote the number of steps within a token interval for ODE integration. As summarized in Table~\ref{tab:cost_compare}, our LASS-ODE method pays a one-time cost to generate $(A_{jk},\boldsymbol{b}_{jk})$ and then evaluates the inexpensive affine field $A_{jk}\boldsymbol{z}+\boldsymbol{b}_{jk}$ at each step, whereas the conditioned Latent ODE must repeatedly evaluate an MLP on $[\boldsymbol{z};\tilde{E}_{\text{final}}^{(jk)}]$. Consequently, the latter has a much higher cost as $n_{\text{step}}$ increases.


\begin{table}[ht]
\centering
\caption{ODE integration complexity per token interval. 
$C_{\text{param}}:= d_{\text{model}}K_{\text{width}} + (L-2)K_{\text{width}}^2 + K_{\text{width}}(d_z^2+d_z)$ is the one-time cost to compute $(A_{jk},\boldsymbol{b}_{jk})$ from $\tilde{E}_{\text{final}}^{(jk)}$ via $f_{\text{param}}$. $C_{\text{lin}}:= d_z^2$ is one evaluation of $A_{jk}\boldsymbol{z}+\boldsymbol{b}_{jk}$. 
$C_{\text{mlp}}:= (d_z+d_{\text{model}})K_{\text{width}} + (L-2)K_{\text{width}}^2 + K_{\text{width}}d_z$ is one evaluation of the conditioned MLP $h(\boldsymbol{z};\tilde{E}_{\text{final}}^{(jk)})$ with input $[\boldsymbol{z};\tilde{E}_{\text{final}}^{(jk)}]$.}
\label{tab:cost_compare}
\small
\begin{tabular}{l l}
\toprule
\textbf{Derivative} & \textbf{Total cost over $n_{\text{step}}$} \\
\midrule
LASS-ODE: $A_{jk}\boldsymbol{z}+\boldsymbol{b}_{jk}$ 
& $\mathcal{O}\!\left(C_{\text{param}} + n_{\text{step}}\,C_{\text{lin}}\right)$ \\
Latent ODE: $h(\boldsymbol{z};\tilde{E}_{\text{final}}^{(jk)})$ 
& $\mathcal{O}\!\left(n_{\text{step}}\,C_{\text{mlp}}\right)$ \\
\bottomrule
\end{tabular}
\vspace{-2mm}
\end{table}

\textbf{LoRA-based Fine Tuning}. For rare and challenging systems, we apply LoRA to the weight matrices in $\mathrm{MHA}$ in Eq. \eqref{eqn:MHA} using only a few system-specific trajectories. This yields substantial performance gains (see Section \ref{subsec:lora_fine_tuning}).

\section{Experiments}
\label{sec:numerical}

\subsection{Settings}
\textbf{Datasets}. We evaluate LASS-ODE on a curated dataset across diverse real-world ODE systems. Dataset descriptions, system portions, and representative training trajectories are provided in Appendix~\ref{app:train-examples}. \textbf{Model and environment} are described in Appendix \ref{app:lassode-params}. \textbf{Benchmark Methods}. We test \textbf{four groups of baselines}: foundation time-series models, classic trainable Transformers for dynamic predictions, Neural ODE families with guaranteed ODE outputs, and hybrid transformer-Neural ODE models. More details are in Appendix \ref{app_baseline}. We exclude traditional system identification, such as SINDy \citep{brunton2016discovering}, since they require system-specific estimation and are not comparable under a multi-system setting. \textbf{Task and Evaluation Metric}. As defined in Section \ref{sec:preliminaries}, the task is to predict the full trajectory conditioned on an observed prefix. Hence, we utilize the prefix ratio, i.e., the first 
30\%, 60\%, or 90\% of the trajectory, as input observations and mean squared error (MSE) to evaluate all methods. 


\begin{table*}[ht]
\caption{
Test MSE ($\times 10^{-2}$) for in-domain and zero-shot ODE systems.
}
\label{tab:in-domain-MSE-Samples}
\centering
\scriptsize
\setlength{\tabcolsep}{3pt}
\begin{tabular}{cc|*{10}{c}}
\toprule
\multirow{2}{*}[0.4em]{ODE Systems} & 
\multirow{2}{*}[0.4em]{Prefix ratio} & 
\makecell{\textbf{LASS-ODE}\\(Ours)} &
\makecell{\textbf{TimesFM}\\\citeyearpar{das2024decoder}} & 
\makecell{\textbf{Chronos}\\\citeyearpar{ansari2024chronos}} & 
\makecell{\textbf{Timer-XL}\\\citeyearpar{liu2024timer}} &
\makecell{\textbf{Informer}\\\citeyearpar{zhou2021informer}} &
\makecell{\textbf{Autoformer}\\\citeyearpar{wu2021autoformer}} &
\makecell{\textbf{Latent ODE}\\\citeyearpar{rubanova2019latent}} &
\makecell{\textbf{Latent MoS}\\\citeyearpar{li2025latent}} &
\makecell{\textbf{ContiFormer}\\\citeyearpar{chen2023contiformer}} \\
\midrule
\multicolumn{11}{c}{In-domain Test}\\
\midrule
\input{in_domain.tex}
\end{tabular}
 \vspace{-2.5mm}
\end{table*}

\subsection{In-domain Prediction}
\label{sec:in_domain_accuracy}
In this subsection, we evaluate in-domain ODE systems whose types appear in the training set, but under unseen initial conditions and physical parameters. The top part of Table~\ref{tab:in-domain-MSE-Samples} reports the MSE on representative systems, and additional results and visualizations across a broader set of ODE systems are provided in Table~\ref{tab:in-domain-MSE-app}, and Figs. \ref{fig:learning_sys_convergence}-\ref{fig:in-domain-app4}. Across diverse datasets, LASS-ODE achieves a more than $70\%$ reduction in error compared to prior methods. Time-series foundation models are optimized for generic sequence regularities and often fail to capture distinctive dynamical behaviors such as non-periodicity, sharp spikes, and irregular patterns (see Figs.~\ref{fig:in-domain-app}-\ref{fig:in-domain-app4}). Other Transformers perform better due to an additional supervised training stage on in-domain data, but are still much worse than LASS-ODE because they do not explicitly model the underlying dynamical laws. Latent ODE and Latent MoS achieve relatively better performance, underscoring the importance of ODE-structured outputs. Finally, ContiFormer computes continuous attention based on ODE flows of keys and values. However, it does not guarantee an ODE-structured output, and its training incurs substantial computational costs.


\subsection{Zero-shot Generalization}
\label{sec:zero_shot_accuracy}

\begin{figure*}[ht]
    \centering
    \includegraphics[width=0.98\linewidth]{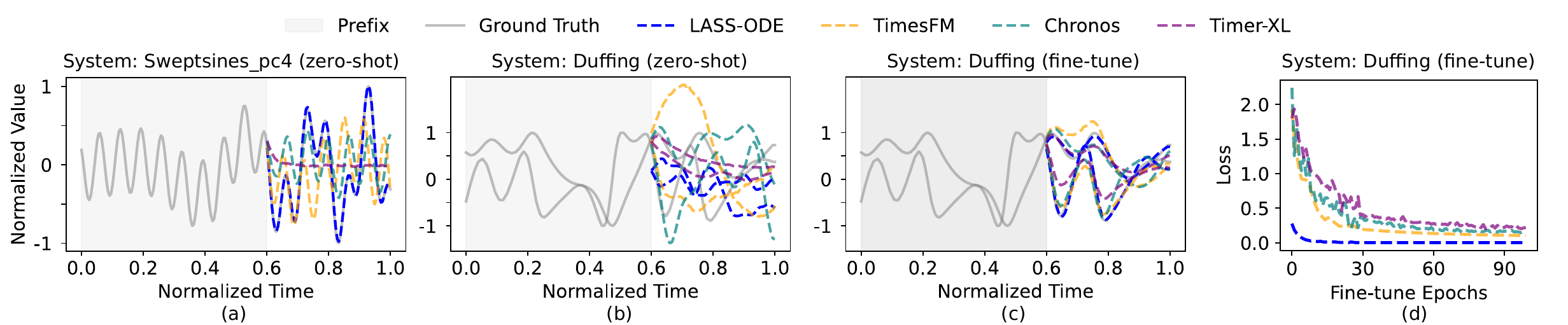}
    \caption{
    (a). Good zero-shot result.
    (b). Bad zero-shot result.
    (c). Fine-tuned result.
    (d). Loss curve in fine-tuning.
    }
    \label{fig:1-in-4}
\end{figure*}

We evaluate zero-shot generalization by testing all models on unseen systems. The bottom of Table \ref{tab:in-domain-MSE-Samples} shows MSE for selected sample systems, and  additional results are in Table \ref{tab:zero-shot-MSE-app} (Appendix \ref{sec:app:zero-shot}). LASS-ODE achieves the lowest error on most systems. For example, Fig. \ref{fig:1-in-4} (a) shows an accurate prediction for cylinder vibration dynamics (Sweptsines\_pc4). For a small number of challenging systems, e.g., the Duffing system in Fig. \ref{fig:1-in-4} (b), all methods perform badly.

\subsection{LoRA-based Fine-tuning}
\label{subsec:lora_fine_tuning}

For systems with poor zero-shot performance, we further evaluate fine-tuning. Specifically, we fine-tune all foundation models on the same training data in an ODE system and test on new trajectories from the same ODE type with unseen initial conditions and ODE parameters. Figs.~\ref{fig:1-in-4}(b) and (c) compare Duffing results before and after fine-tuning. LASS-ODE adapts rapidly and achieves the most accurate predictions, consistent with loss curves in Fig.~\ref{fig:1-in-4}(d). Table~\ref{tab:fine-tune-MSE-Samples} reports the test MSE. More results are in Figs. \ref{fig:app-fine-tuned} and \ref{fig:app-fine-tuned-2}.


\subsection{CSH Effectiveness}
\label{subsec:csh_effectiveness}
\begin{figure}[ht]
    \centering
    \includegraphics[width=1\linewidth]{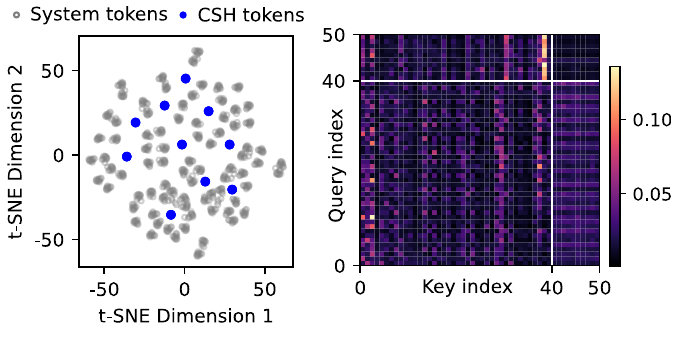}
    \caption{The t-SNE visualization and heatmap.} 
    \label{fig:heatmap}
\end{figure}

To probe why the CSH is effective, we visualize (i) a t-SNE projection \citep{maaten2008visualizing} of the learned CSH tokens together with token embeddings from 15 systems, and (ii) the first-layer attention map computed over the concatenation of CSH and system-specific token embeddings. Fig.~\ref{fig:heatmap} shows the results, which jointly reveal the proximity between CSH and system-specific tokens, their mutual information exchange, and the update mechanism of the CSH. The t-SNE plot suggests that training drives CSH tokens toward the local centers of a group of tokens, consistent with their role as shared prototypes that encode common ODE structure. The bottom-right block of the attention heatmap indicates that system-specific token queries have non-trivial attention scores on CSH tokens and extract the CSH information. The top-left and top-right blocks indicate that, for CSH token queries, the update is dominated by system-specific tokens, while interactions among CSH tokens are very weak. This suggests that LASS-ODE learns to update each CSH token primarily by aggregating evidence from system-specific data, rather than mixing with other CSH tokens, consistent with the goal that different CSH tokens specialize in distinct ODE patterns.

\begin{table}[h]
\caption{
Test MSE ($\times 10^{-2}$) for fine-tuned systems.
}
\label{tab:fine-tune-MSE-Samples}
\centering
\scriptsize
\setlength{\tabcolsep}{3pt}
\begin{tabular}{lc|*{4}{c}}
\toprule
\multirow{2}{*}[0.4em]{ODE Systems} & 
\multirow{2}{*}[0.4em]{Prefix ratio} & 
\makecell{\textbf{LASS-ODE}\\(Ours)} &
\makecell{\textbf{TimesFM}\\\citeyearpar{das2024decoder}} & 
\makecell{\textbf{Chronos}\\\citeyearpar{ansari2024chronos}} & 
\makecell{\textbf{Timer-XL}\\\citeyearpar{liu2024timer}} \\
\midrule
\input{fine-tune.tex}
\end{tabular}
\end{table}

\subsection{Ablation Study}
\label{subsec:ablation}
In this subsection, we fix the prefix ratio at 60\% and report the average in-domain test MSE across all systems. Following the pipeline in Fig.~\ref{fig:bigpic} (left to right), we evaluate variants that: $(1)$ use channel-dependent processing; $(2)$ replace GRU and time modulation with an MLP encoder, $(3)$ use Fourier feature (FF) time encoding; $(4)$ use rotary positional embeddings (RoPE) for time encoding; $(5)$ remove channel encoding; $(6)$ remove the CSH module; $(7)$ replace MoE with a single MLP; and $(8)$ replace the piecewise-linear (PWL) derivative in Eq.~\eqref{eq:piecewise_odesolve} with a nonlinear MLP. 

\begin{figure}[h]
    \centering
    \begin{minipage}[t]{0.58\linewidth}
        \centering
        \includegraphics[height=2.7cm]{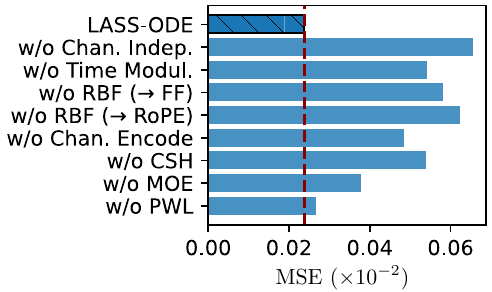}
        \caption{Ablation results.}
        \label{fig:ablation}
    \end{minipage}
    \hfill
    \begin{minipage}[t]{0.38\linewidth}
        \centering
        \includegraphics[height=2.7cm]{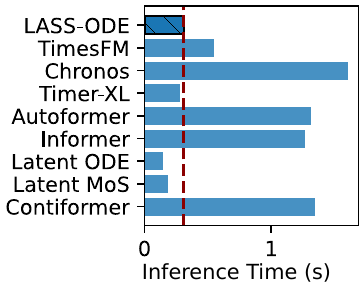}
        \caption{Inference time.}
        \label{fig:inference}
    \end{minipage}
\end{figure}

Fig. \ref{fig:ablation} illustrates the results. Enforcing channel dependence at the encoder input increases error by more than $3\times$, suggesting that early coupling is sensitive to channel heterogeneity and can degrade learning. The GRU encoder and time modulation are also crucial for temporally consistent token embeddings: replacing them with local MLP-based embeddings increases the error by about $2\times$. Replacing the RBF time encoding with FF or RoPE yields a $2\!\sim\!3\times$ error increase, as RBF can emphasize each token's \emph{absolute} normalized start time $t_k^{\text{start}}\!\in[0,1]$ (see Section~\ref{subsec:ode_tokenization}) via multiple kernels, whereas FF and RoPE primarily capture \emph{relative} positional information. Channel encoding further helps distinguish channels within each system and prevents ambiguity in subsequent attention layers, and removing it increases error by about $2\times$. Removing the CSH or MoE leads to a $1.5\times\!\sim\!2.2\times$ degradation, indicating that sparse experts capture diverse ODE patterns while the CSH aggregates transferable structure across systems (we provide additional sensitivity analysis below). Finally, replacing the piecewise-linear derivative in $\mathrm{ODESolve}$ with a nonlinear MLP slightly worsens accuracy and increases training time by more than $15\times$, confirming that our linearization is accurate and efficient.

\subsection{Inference Time}

Fig.~\ref{fig:inference} reports the inference time for predicting $16$ trajectories across different ODE systems. Thanks to the sparse MoE and the PWL ODE decoder, LASS-ODE achieves moderate latency, capable of real-time ODE prediction.

\subsection{Sensitivity Analysis}
\label{subsec:sensitivity_analysis}

\begin{figure}[h]
    \centering
    \includegraphics[width=1.0\linewidth]{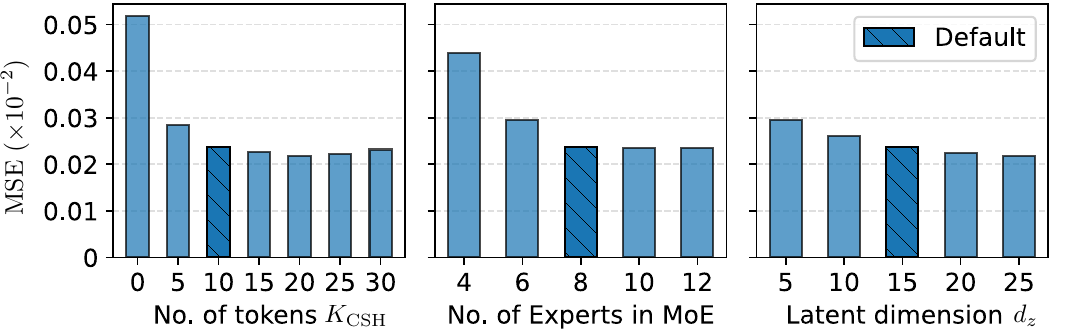}
    \caption{Sensitivity analysis results.}
    \label{fig:sensitivity}
\end{figure}

We use the ablation setup for sensitivity analysis, varying $K_{\text{CSH}}$, the MoE expert count, and the latent dimension $d_z$ of $\boldsymbol{z}$ in Eq.~\eqref{eq:piecewise_odesolve}, which together control model capacity. Fig.~\ref{fig:sensitivity} summarizes the results. The test MSE generally decreases as we increase these hyperparameters. Beyond a moderate point ($K_{\text{CSH}}\!\ge\!10$, \#experts $\!\ge\!8$, and $d_z\!\ge\!15$), performance saturates and LASS-ODE attains its lowest test MSE. We attribute this to the finite diversity of local ODE structures in the current training dataset. With a more diverse and informative ODE dataset, we expect the saturation point to shift right and the test MSE to decrease further.

\section{Related Work}
\label{sec:related}

\textbf{Transformer-based Time Series.}
Large pretraining has recently enabled time-series models with cross-domain generalization. TimeGPT introduced a GPT-style decoder \citep{garza2023timegpt}, and Lag-Llama scaled decoder-only transformers with lagged covariates for probabilistic prediction \citep{lagllama}. Moirai unified training across frequencies and variable sets with any-variate attention \citep{liu2024moirai}. Chronos and TimesFM leverage scaling-and-quantization or flexible horizon objectives to achieve strong zero-shot accuracy across benchmarks \citep{ansari2024chronos,das2024decoder}. In parallel, a line of Transformers has focused on architectural novelties: TimesNet models multi-periodicity by transforming sequences into multi-scale representations \citep{wu2022timesnet}, iTransformer inverts the typical time-as-token design to better capture cross-variable dependencies \citep{liu2023itransformer}, Timer-XL targets long-context, unified prediction via multivariate next-token prediction \citep{liu2024timer}, and Sundial advances native generative pretraining for time series \citep{liu2025sundial}. Collectively, these models demonstrate the promise of foundation architectures for time series but overlook the capacity to capture dynamics. 

\textbf{Cross-System Knowledge Sharing}. Explicit knowledge sharing can improve model efficiency. Related ideas appear in multi-modal and multi-task learning. Perceiver IO \citep{jaegle2021perceiver} processes multi-modal inputs via cross-attention from inputs to a latent array, followed by self-attention over the latent tokens. Similarly, cross-task attention has been explored through parallel self- and cross-attention modules to enable bi-directional information exchange between tasks \citep{lopes2023cross}. However, these designs (1) primarily instantiate pairwise (one-to-one) cross-attention, which scales poorly as the number of systems grows, and (2) rely on compositions of self- and cross-attention with extra architectural choices. In Section \ref{subsec:intra-inter-attention}, we show simple self-attention is sufficient. An alternative direction is retrieval-augmented modeling \citep{borgeaud2022improving}, which augments predictions by retrieving relevant information from a large external database. However, applying retrieval methods to ODE systems requires additional effort to construct and maintain the database, as well as a nontrivial retrieval cost.

\textbf{Neural ODE/CDE/SDE Families for Learning Dynamics.}
Continuous-time neural architectures offer a complementary line of work. Neural ODEs \citep{chen2018node} introduced black-box differential solvers as network layers, enabling flexible vector-field learning. Extensions such as Latent ODEs and ODE-RNNs \citep{rubanova2019latent} infer latent initial states from irregular observations; Neural CDEs generalize recurrent networks to pathwise control signals \citep{kidger2020neural}; and Neural RDEs extend to rough paths \citep{morrill2020nrde}. These families establish the feasibility of learning dynamics directly in continuous time, but they are typically trained as narrow, system-specific models without cross-domain scalability. Our contribution is to integrate this constrained learning into a foundation model for ODEs, which achieves systematic generalization across different dynamical regimes.

\section{Conclusion, Limitation, and Future Work}
We introduced LASS-ODE, a foundation-model paradigm that connects large-scale pretraining to real-world dynamical systems. While LASS-ODE is developed for ODE simulation, the core ideas extend to broader industrial foundation models, including planning and optimization. Our key insights are two-fold: (1) enforce physical constraints in a scalable manner via token-wise relaxation; and (2) extract cross-system common knowledge through architectural designs that enable shared structure retrieval. We further provide a unified treatment pipeline for heterogeneous dynamical data and identify effective modern modules for end-to-end training. Experiments on diverse ODE datasets show that LASS-ODE achieves the state-of-the-art performance as a general ODE foundation model. This paper only focused on autonomous systems. On the application side, extending LASS-ODE to control tasks is a promising direction. More broadly, we view LASS-ODE as a step toward foundation models that combine the universality of large-scale pretraining with the fidelity of physics-based modeling, including future products such as LASS-SDE, LASS-ODE-Control, and LASS-OPT.

\clearpage

\section*{Impact Statement}
This paper presents work whose goal is to advance the field of machine learning. There are many potential societal consequences of our work, none of which we feel must be specifically highlighted here.

\bibliography{example_paper}
\bibliographystyle{icml2026}

\newpage
\appendix
\onecolumn
\section{Appendix}

\subsection{
Dataset Description}

\label{app:train-examples}

To evaluate LASS-ODE, we combine established nonlinear dynamical systems benchmarks with a large and diverse library of simulated dynamical systems, focusing on systems governed by ordinary differential equations (ODEs).

We first use datasets from the Nonlinear Benchmark repository \cite{nonlinearbenchmark}, which provides standardized real-world and synthetic time-series from nonlinear dynamical systems. From this repository, we include several Wiener-Hammerstein (WH) process benchmarks \cite{schoukens2015parametric, schoukens2017three}: the standard WH system, a Parallel WH variant with multiple branches, and a version with dominant process noise entering before the nonlinearity. These datasets exhibit nonlinear temporal dynamics with memory, internal state evolution, and realistic noise.

We also include datasets from other physical domains to broaden dynamical diversity. The Unsteady Fluid Mechanics benchmark \cite{decuyper2024canonical} is derived from computational fluid dynamics (CFD) simulations of flow past a transversely oscillating cylinder. The dataset contains multiple types of imposed motions (sinesweep, sine, multisine) on the cylinder, and the velocity, pressure, and vorticity of the flow in the wake of the cylinder are measured on a 31 x 31 grid. This dataset introduces high-dimensional, spatially distributed, and strongly nonlinear temporal dynamics.

Beyond these established benchmarks, we construct a large curated library of simulated time-series generated from classical ODE models. The goal of this library is to expose the model to a wide range of dynamical behaviors, regimes, and time scales. This library spans:
\begin{itemize}
    \item Low-dimensional test equations include Bernoulli, Cauchy-Euler, and Emden-Fowler equations.
    \item Nonlinear oscillatory systems include the Van der Pol oscillator (across multiple stiffness parameters), damped and undamped pendulum systems, the Duffing oscillator, and the Ermakov-Pinney equation.
    \item Biological and reaction dynamics include logistic growth, the Lotka-Volterra predator–prey model, the FitzHugh-Nagumo neuron model, and the Hopf normal form.
    \item Chaotic systems include the Lorenz-63 and Rössler attractors, as well as the higher-dimensional Lorenz-96 model.
    \item Power system dynamics include swing-equation and droop-controlled grid frequency models.
\end{itemize}
For each ODE system, we generate multiple trajectories under diverse initial conditions and parameter settings to capture both transient and long-term behaviors. Simulations are sampled at multiple temporal resolutions, ranging from millisecond-scale to second-scale sampling, to encourage robustness across time scales. All trajectories are stored in .CSV format for data loading.

Across all established benchmark datasets and our simulated ODE library, the total data volume used in our experiments is approximately 40\,GB, comprising roughly 500 million time points.

\begin{figure}[h]
    \centering
    \includegraphics[width=0.95\linewidth]{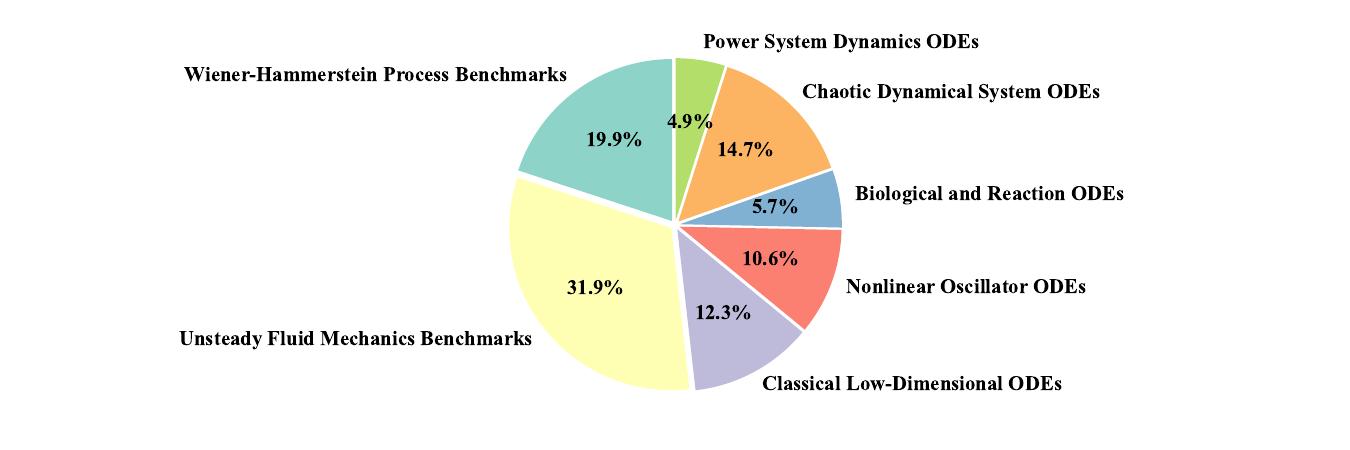}
    \caption{Ratios of data sources used in this work. Example trajectories from the constructed dataset are shown in Fig. \ref{fig:example-traj}.}
    \label{fig:pie-chart}
\end{figure}

\begin{figure}[h]
    \centering
    \includegraphics[width=1\linewidth]{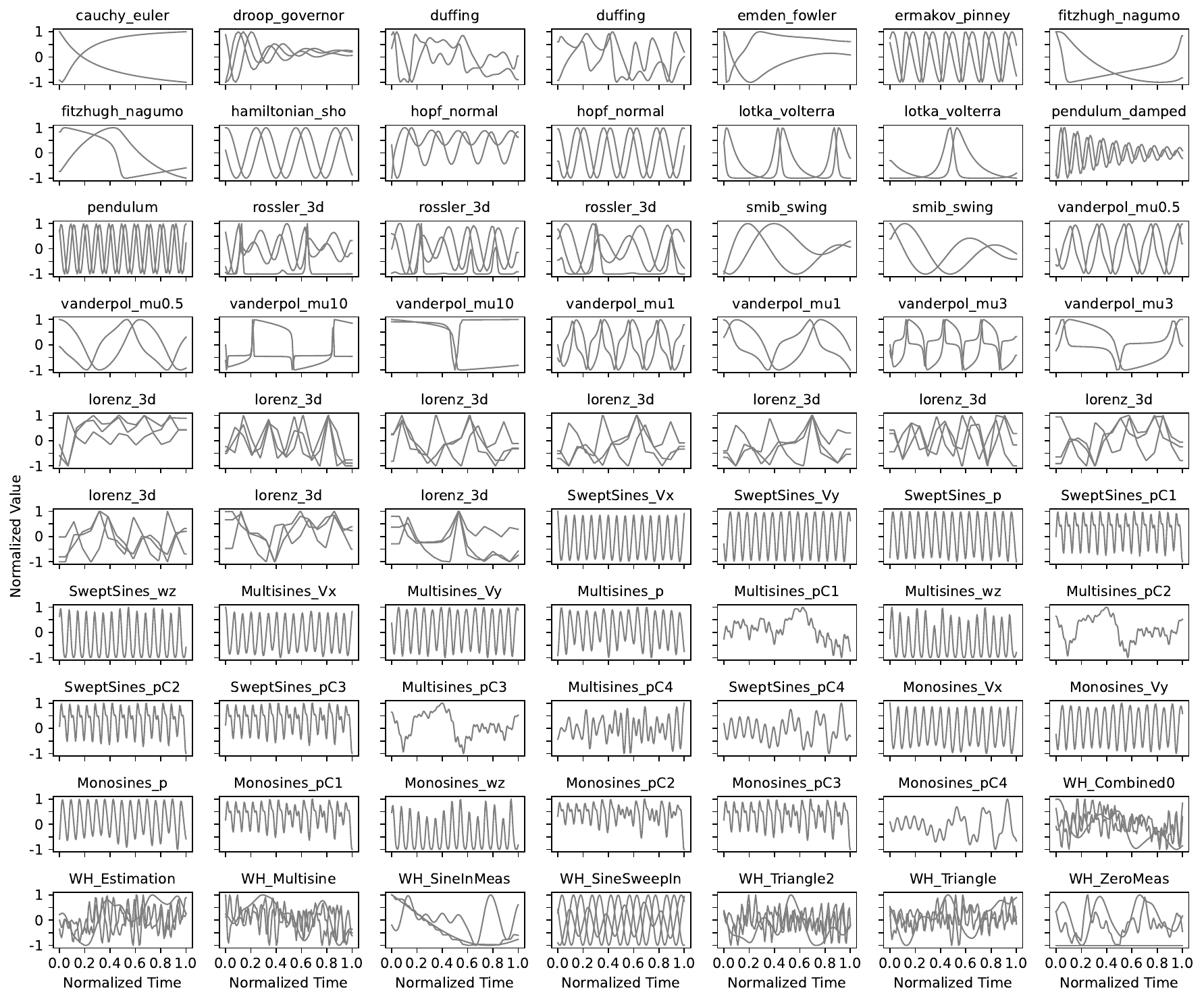}
    \caption{Example trajectories from the constructed dataset, with each panel corresponding to a different dynamical system.}
    \label{fig:example-traj}
\end{figure}

Fig. \ref{fig:example-traj} presents example trajectories from the constructed dataset used in this work. Each panel corresponds to a single dynamical system, and multiple curves within a panel represent different state dimensions of that system. For high-dimensional systems, we divide the state variables across multiple panels, with each panel showing two or three dimensions for visual clarity.

\clearpage
\subsection{Models and Environments}
\label{app:lassode-params}

\paragraph{Environment.} All experiments were conducted on a high-performance computing cluster equipped with $4$ NVIDIA A100 GPUs, each with 80 GB of GPU memory.

\paragraph{Hyperparameters (default run).}
Unless otherwise stated, our experiments use the following configuration:
\begin{table}[h!]
\centering
\caption{Model and training hyperparameters.}
\small
\begin{tabular}{l l l}
\toprule
\textbf{Category} & \textbf{Name} & \textbf{Value} \\
\midrule
\midrule

Data size         & ODE time points & 50-200 \\
& State dimension $d_x$ & 1-5\\
& Maximal state dimension $d_x^{\text{max}}$ & 10\\
& Number of prefixes & 56 \\
& Prefix ratio & 0.3-0.9\\
Architecture & Latent dim $d_z$ & 15 \\
             & Embed dim $d_{\text{model}}$& 256 \\
             & GRU hidden & 256 \\
             & GRU layers & 2 \\
             & Number of kernels in RBF embedding $K_{\text{rbf}}$ & 64 \\
             &$\sigma$ in RBF embedding & 0.25\\
             & Number of tokens per trajectory $K_{\text{token}}$ & 40 \\
             & Number of global tokens in CSH: $K_{\text{CSH}}$ & 10 \\
             & Attention heads & 8 \\
             & Transformer depth $L$ & 6 \\
             & Number of experts in MoE & 8\\
             & Number of activated experts & 2\\
             & Number of hidden units & 512 \\
             & ODE solver & \texttt{rk4} \\
\addlinespace
Training & Optimizer & AdamW \\
         & Learning rate & $1\times 10^{-3}$ \\
         & Betas & $(0.9,\;0.95)$ \\
         & Weight decay & 0.05 \\
         & Batch size & 16 \\
         & Epochs & 1500 \\
\addlinespace
Finetuning & Method & LoRA \\
           & Learning rate & $5\times 10^{-3}$ \\
           & Epochs & 30-100 \\
\bottomrule
\end{tabular}
\end{table}

\paragraph{Model Size.} In general, LASS-ODE has 15M parameters, which is sufficient for current tests. We plan to scale the model size and training data size in the future.

\paragraph{Model Details.}

\textbf{GRU Encoder}. We use 2 layers GRUs with a hidden dimension to be 256 to process each channel's data. The input dimension is $2$ (each channel's data and the time $\Delta t$). A complete trajectory's hidden state is computed at one time, and sliced according to the current prefix ratio (0.3-0.9). So, the encoding computation is efficient.

\textbf{RBF-based Time Modulation}. We employ 64 kernels to compute the RBF embedding for each token's start time $t_k^{\text{start}}$ and use $\sigma = 0.25$ to scale the kernel. For the time modulation, as shown in Eq. \eqref{eq:modulation}, we set the hidden unit of the two MLPs to be $512$, and the activation function to be Tanh. They both have 2 layers. Their output dimensions are 512 (the first MLP that outputs $\boldsymbol{\gamma}_k$ and $\boldsymbol{\beta}_k$) and 256 (the second MLP), respectively. 

\textbf{Channel Encoding}. We set the learnable lookup table $C\in\mathbb{R}^{10\times 256}$.

\textbf{CSH}. We set $E_{\text{CSH}}\in\mathbb{R}^{10\times 256}$. 

\textbf{Transformer backbone module}. The input dimension of the Transformer backbone depends on the ODE dimensions (i.e., the number of channels). We set the number of tokens for each channel to be $40$. For example, if the input channel $d_x=2$, the number of tokens from the system is $40\times 2 = 80$. The total input embedding matrix $E_0$ in Eq. \eqref{eq:inter_intra_attention} has the size $90\times 256$. In the $\mathrm{MHA}$, we set $8$ heads. For the following MoE, we have $8$ experts, $2$ of them are activated each time. For each expert, it's an MLP with 2 layers, activated by GELU, and with 512 hidden units. Then, we stack $L=6$ blocks. 

\textbf{Parameter Estimator}. In Eq. \eqref{eq:param_estimator}, we use an MLP $f_{\text{param}}$ to decode the embeddings to estimate the linear ODE parameter. The MLP has 2 layers with Tanh as the activation function. The input dimension is $256$, the hidden unit is $256$, and the output dimension is $d_z^2 + d_z= 240$. 

\textbf{ODE Solver}. We use rk4 to solve the ODE and all other parameters are by default in \texttt{torchdiffeq.odeint} \citep{chen2018node}. 

\textbf{Linear Decoder}. As shown in the last equation of Eq. \eqref{eq:piecewise_odesolve}, the linear decoder has the size $W_{\text{dec}}\in\mathbb{R}^{15\times 1}$ and $b_{\text{dec}}\in\mathbb{R}$. 

\clearpage
\subsection{Baseline Models}
\label{app_baseline}
We test four types of models: (1) foundation time-series models, including TimesFM \citep{das2024decoder}, TimeGPT \citep{garza2023timegpt}, Chronos \citep{ansari2024chronos}, and Timer-XL \citep{liu2024timer}, evaluated primarily under zero-shot inference (and fine-tuning when supported). (2) Classic trainable Transformers for dynamic predictions, including Informer \citep{zhou2021informer}, Autoformer \citep{wu2021autoformer}. (3) Neural ODE–based models that produce ODE-structured trajectories, including Latent ODE \citep{rubanova2019latent} and Latent MoS \citep{li2025latent}. (4) Hybrid Transformer–ODE architectures that combine a Transformer backbone with neural ODE–structured outputs, such as ContiFormer \citep{chen2023contiformer}.  We exclude traditional system identification, such as SINDy \citep{brunton2016discovering}, since they require system-specific estimation and are not comparable under a multi-system setting.

\clearpage
\subsection{
Convergence of LASS-ODE to Ground-Truth ODE Trajectories Over Training
}
\label{sec:app:convergence}

\begin{figure*}[ht]
    \centering
    \includegraphics[width=1\linewidth]{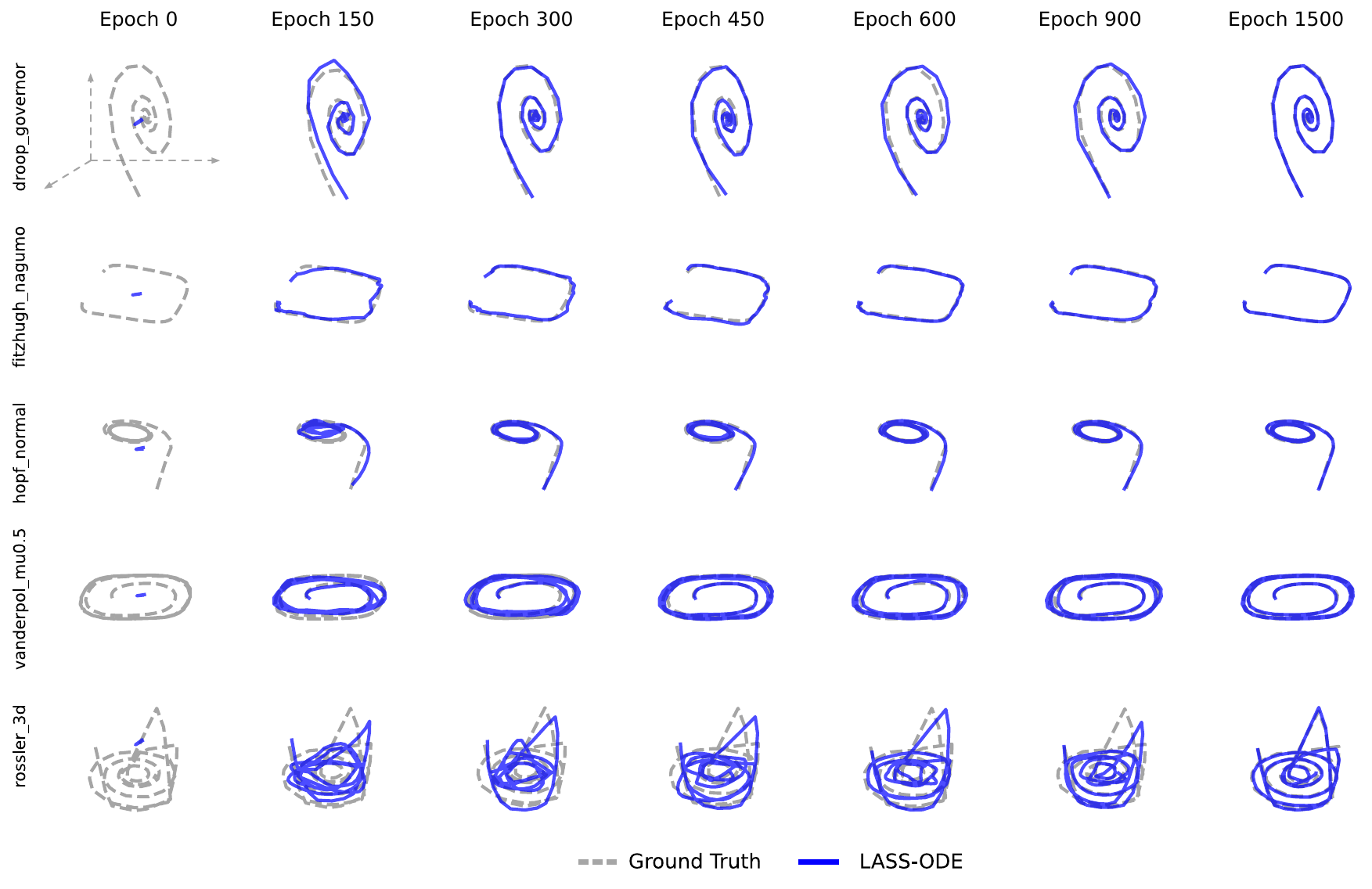}
    \caption{
    Progressive learning of ODE trajectories by LASS-ODE across training epochs. Each row corresponds to a different ODE system from the training dataset, and each column shows model predictions at increasing training epochs (0, 150, 300, 450, 600, 900, and 1500). In each panel, the ground-truth trajectory is plotted together with the trajectory reconstructed by LASS-ODE.
    }
    \label{fig:learning_sys_convergence}
\end{figure*}

This section provides a qualitative visualization of how LASS-ODE progressively learns the underlying dynamics of ODE-governed time-series during training. Each row in Fig. \ref{fig:learning_sys_convergence} corresponds to a distinct dynamical system from the training set, while columns show model predictions at increasing training epochs (0, 150, 300, 450, 600, 900, and 1500). Within each panel, the ground-truth trajectory is plotted alongside the trajectory reconstructed by LASS-ODE. At early epochs (0, 150, and 300), the predicted trajectories exhibit noticeable geometric and phase mismatches, reflecting the model’s limited understanding of the governing dynamics at this stage. As training proceeds, the predicted curves gradually capture the global shape, local curvature, and long-term evolution patterns of the true trajectories. By later epochs (900 and 1500), LASS-ODE produces trajectories that closely align with the ground truth across a variety of dynamical regimes, indicating that the model has internalized system-level structural information rather than merely fitting short-term signal patterns.

\clearpage
\subsection{
Results of In-Domain Test Trajectories and MSE}

\label{sec:app:in-domian}

\begin{figure}[ht]
    \centering
    \includegraphics[width=0.78\linewidth]{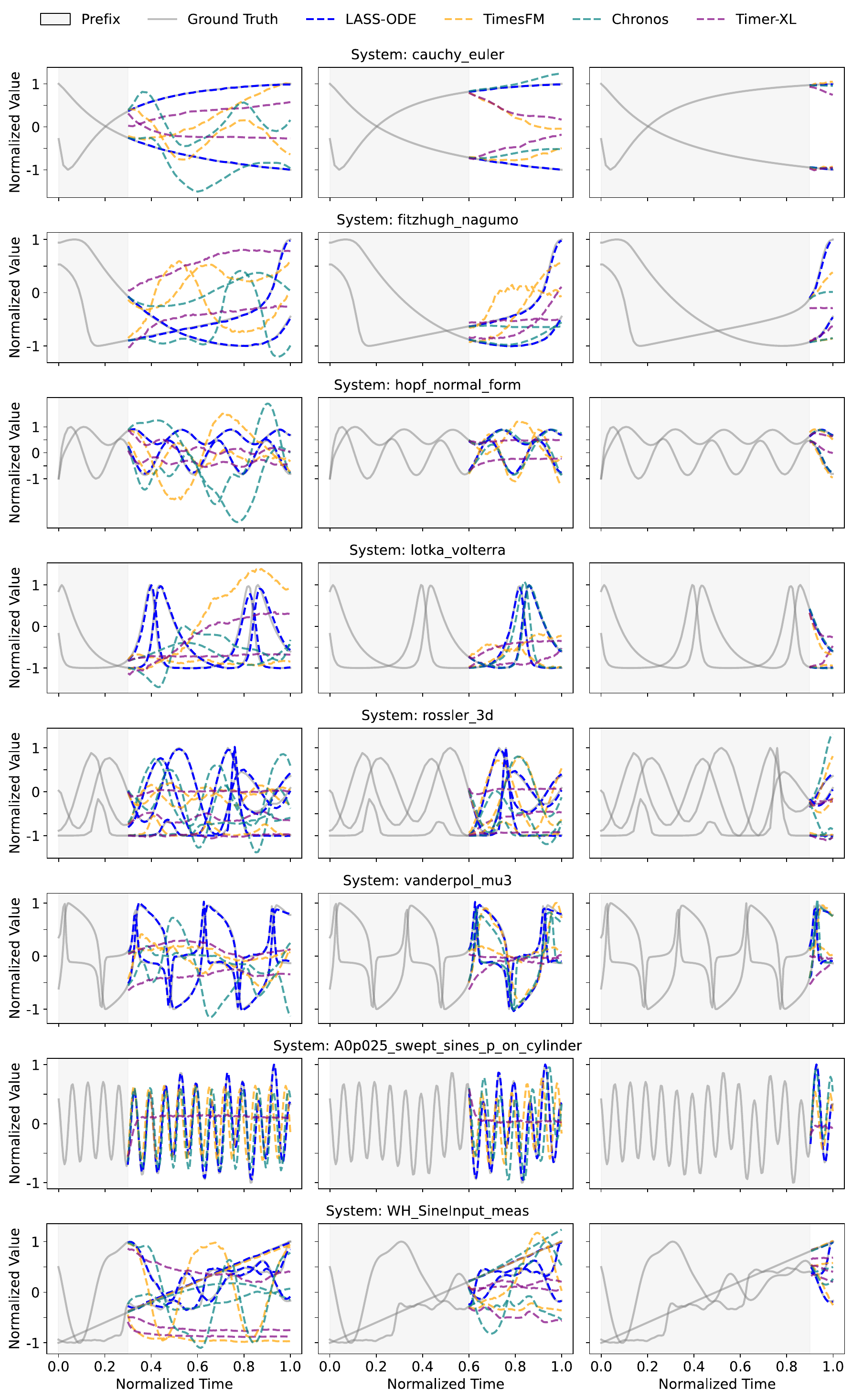}
    \caption{Examples of trajectories generated by time-series foundational models on the in-domain test set.}
    \label{fig:in-domain-app}
\end{figure}

\begin{figure}[ht]
    \centering
    \includegraphics[width=0.78\linewidth]{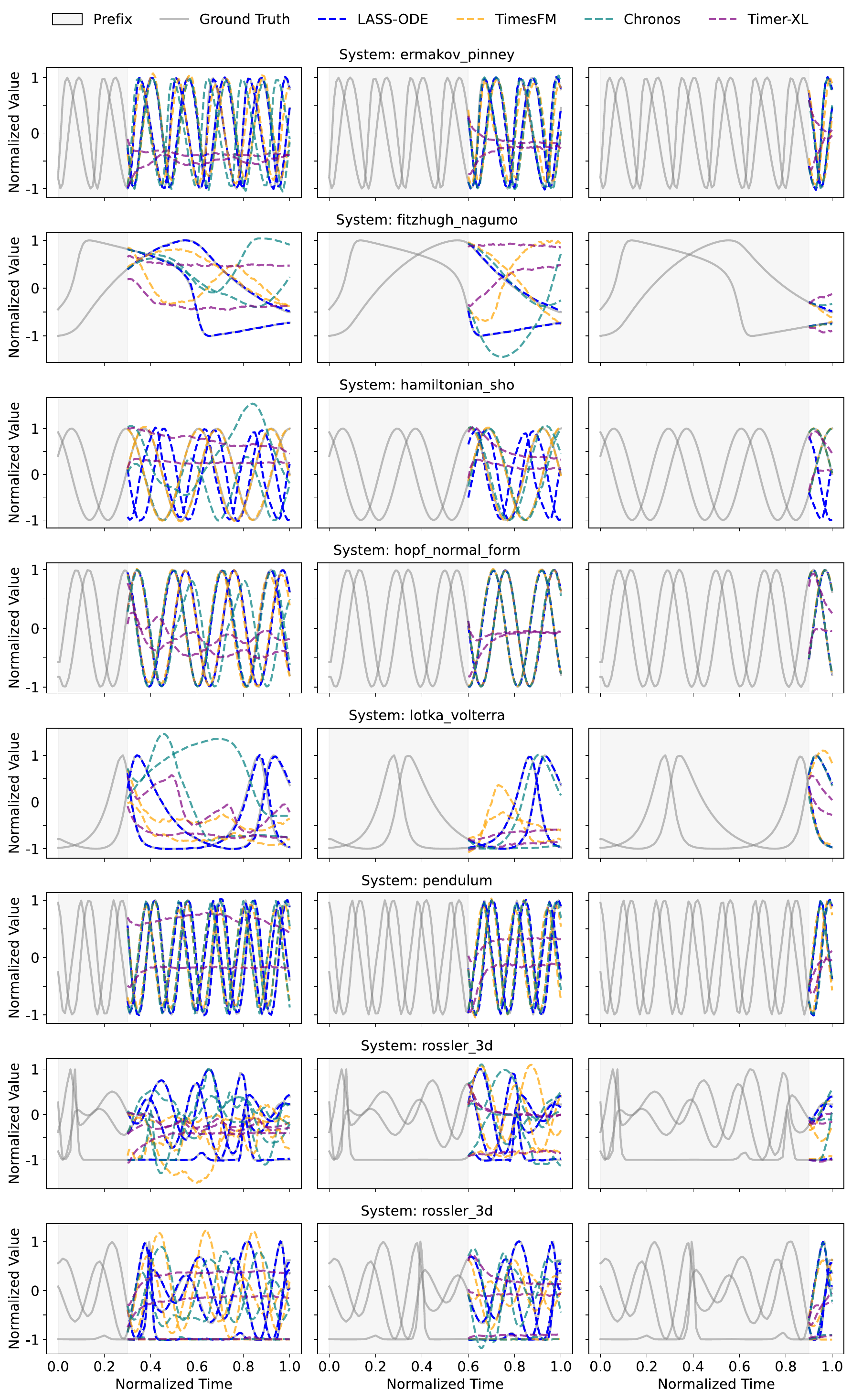}
    \caption{Examples of trajectories generated by time-series foundational models on the in-domain test set.}
    \label{fig:in-domain-app1}
\end{figure}

\begin{figure}[ht]
    \centering
    \includegraphics[width=0.78\linewidth]{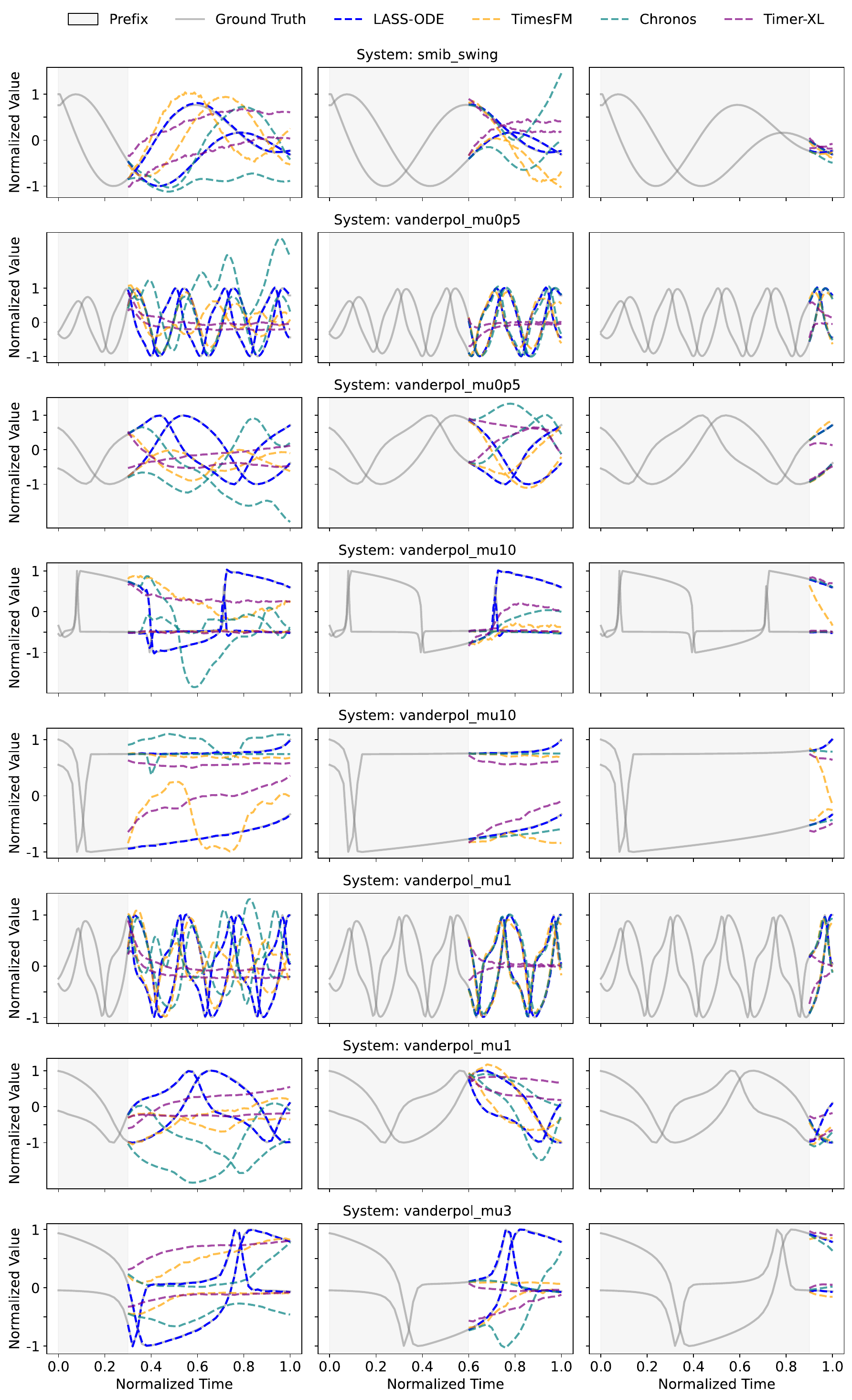}
    \caption{Examples of trajectories generated by time-series foundational models on the in-domain test set.}
    \label{fig:in-domain-app2}
\end{figure}

\begin{figure}[ht]
    \centering
    \includegraphics[width=0.78\linewidth]{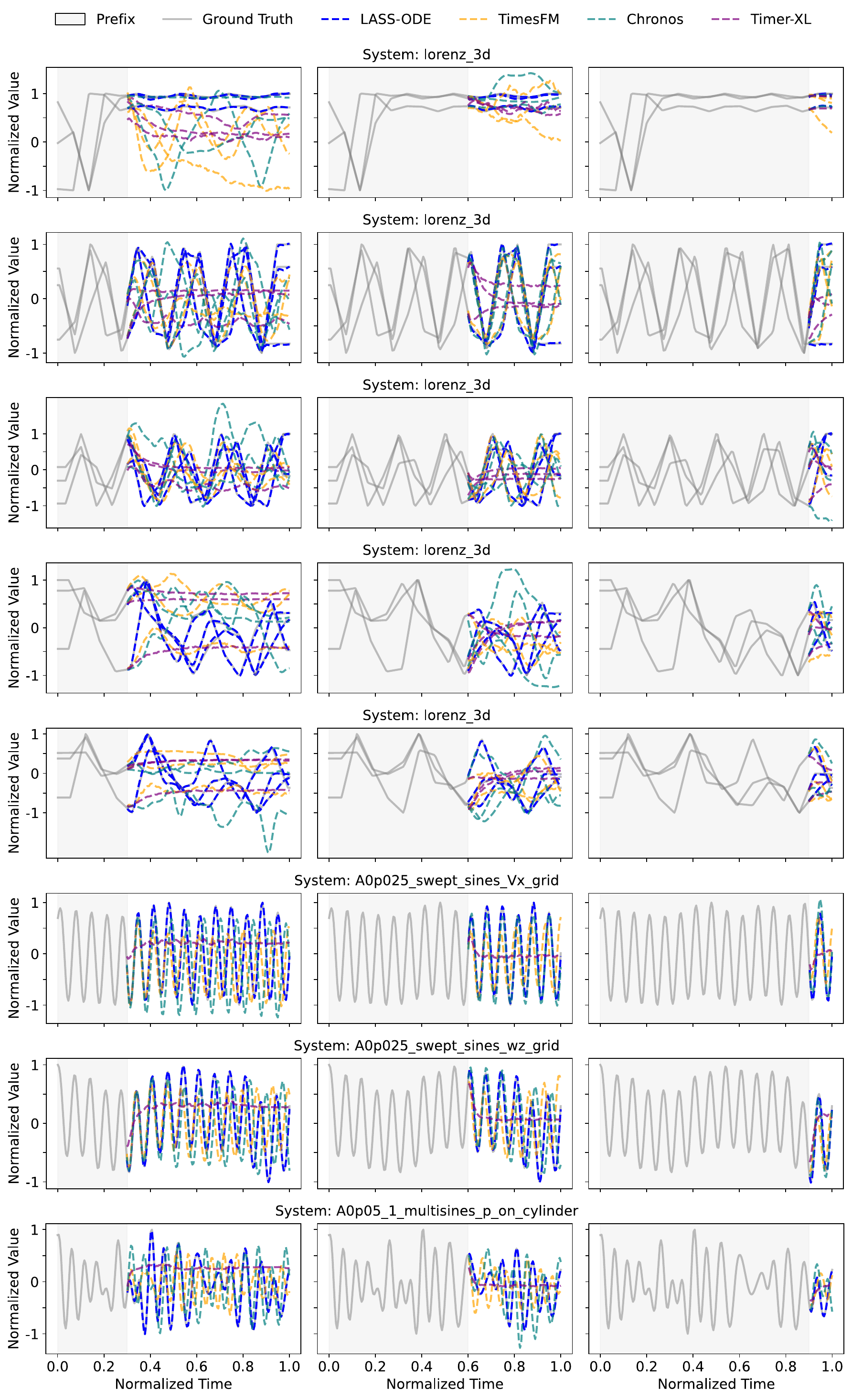}
    \caption{Examples of trajectories generated by time-series foundational models on the in-domain test set.}
    \label{fig:in-domain-app3}
\end{figure}

\begin{figure}[ht]
    \centering
    \includegraphics[width=0.78\linewidth]{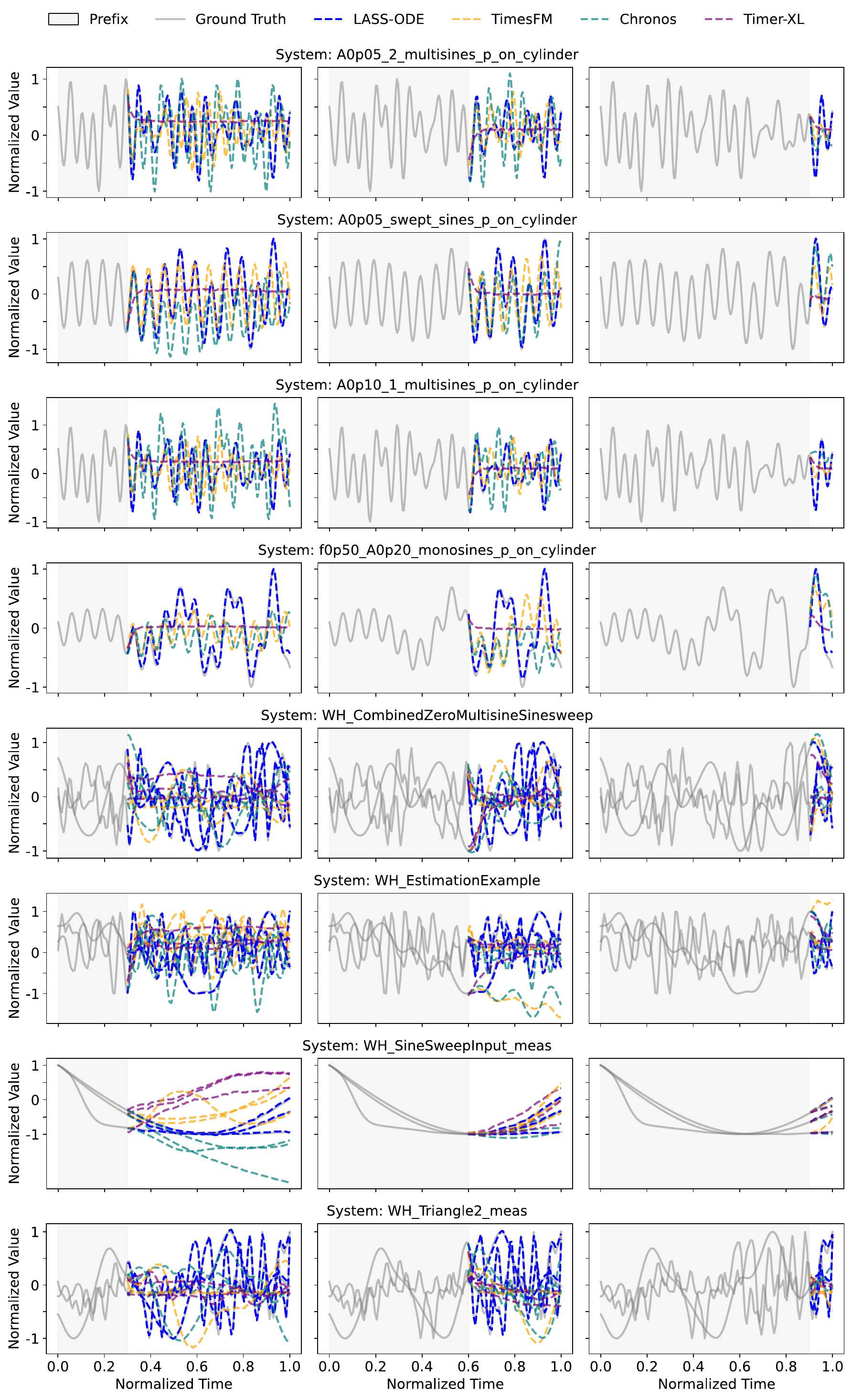}
    \caption{Examples of trajectories generated by time-series foundational models on the in-domain test set.}
    \label{fig:in-domain-app4}
\end{figure}

We show the table of MSE of in-domain test in Table \ref{tab:in-domain-MSE-app}.
We show the trajectory comparison of us and time-series
foundation models in Figs. \ref{fig:in-domain-app}-\ref{fig:in-domain-app4}. The prefix ratio is $30\%$, $60\%$, and $90\%$.

\begin{table*}[ht]
\caption{
Test MSE ($\times 10^{-2}$) for in-domain systems.
}
\label{tab:in-domain-MSE-app}
\centering
\tiny
\begin{tabular}{p{1.4cm}*{27}{p{0.11cm}}}
\toprule
\multirow{2}{*}[0.4em]{Models} & 
\multicolumn{3}{c}{\makecell{\textbf{LASS-ODE}\\(Ours)}} &
\multicolumn{3}{c}{\makecell{\textbf{TimesFM}\\\citeyearpar{das2024decoder}}} & 
\multicolumn{3}{c}{\makecell{\textbf{Chronos}\\\citeyearpar{ansari2024chronos}}} & 
\multicolumn{3}{c}{\makecell{\textbf{Timer-XL}\\\citeyearpar{liu2024timer}}} &
\multicolumn{3}{c}{\makecell{\textbf{Informer}\\\citeyearpar{zhou2021informer}}} &
\multicolumn{3}{c}{\makecell{\textbf{Autoformer}\\\citeyearpar{wu2021autoformer}}} &
\multicolumn{3}{c}{\makecell{\textbf{Latent ODE}\\\citeyearpar{rubanova2019latent}}} &
\multicolumn{3}{c}{\makecell{\textbf{Latent MoS}\\\citeyearpar{li2025latent}}} &
\multicolumn{3}{c}{\makecell{\textbf{ContiFormer}\\\citeyearpar{chen2023contiformer}}} \\

\cmidrule(lr){2-4}\cmidrule(lr){5-7}\cmidrule(lr){8-10}\cmidrule(lr){11-13}\cmidrule(lr){14-16}\cmidrule(lr){17-19}\cmidrule(lr){20-22}\cmidrule(lr){23-25}\cmidrule(lr){26-28}

Prefix ratio &
30\% & 60\% & 90\% &
30\% & 60\% & 90\% &
30\% & 60\% & 90\% &
30\% & 60\% & 90\% &
30\% & 60\% & 90\% &
30\% & 60\% & 90\% &
30\% & 60\% & 90\% &
30\% & 60\% & 90\% &
30\% & 60\% & 90\% \\
\midrule
\input{in_domain_app.tex}
\end{tabular}
\end{table*}

\clearpage
\subsection{
Results of Zero-Shot ODE Results}

\label{sec:app:zero-shot}

\begin{figure}[ht]
    \centering
    \includegraphics[width=0.45\linewidth]{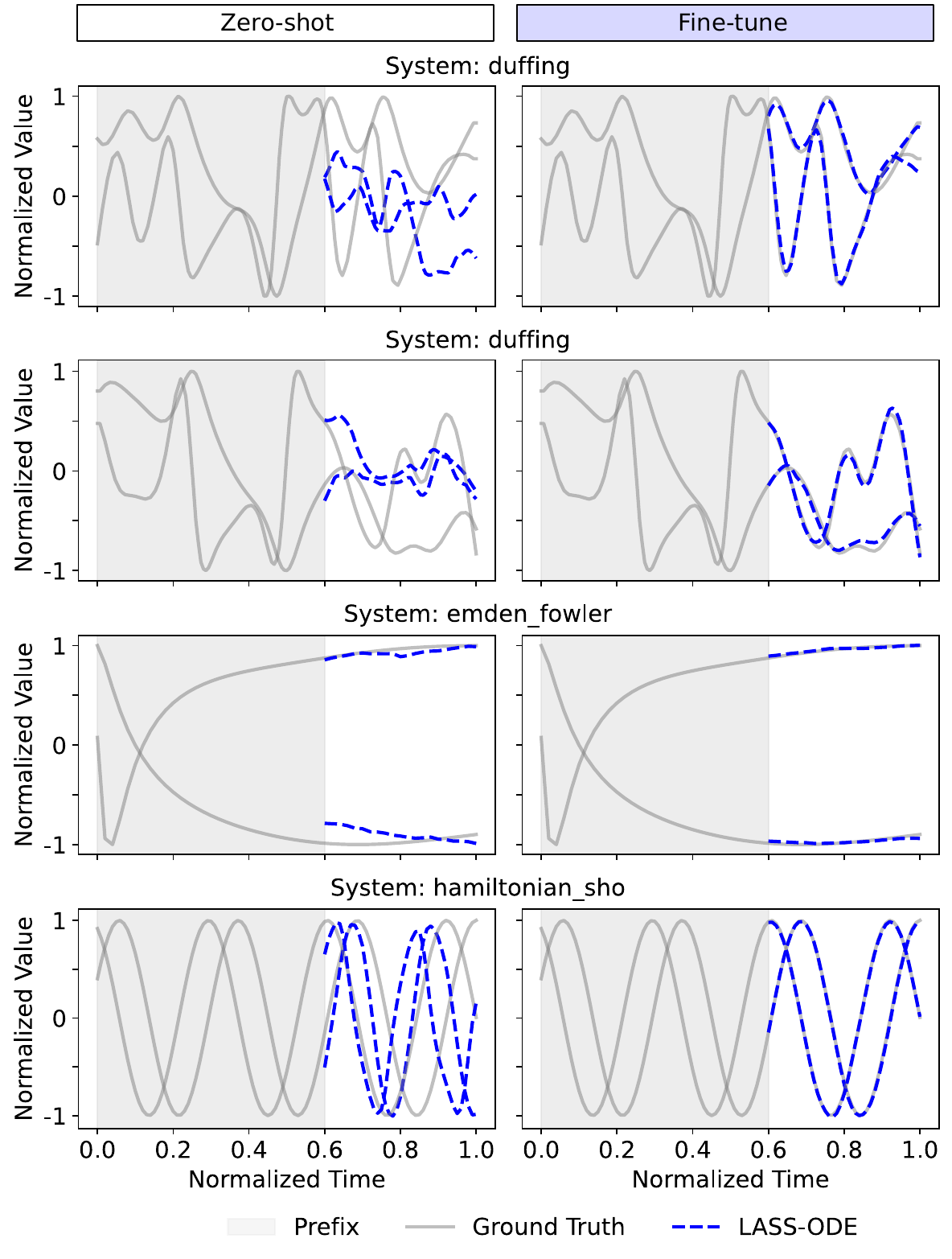}
    \caption{Performance comparison of the same base system under zero-shot inference (left) and after LoRA fine-tuning (right) for 30 epochs. The prefix ratio is set to $60\%$.}
    \label{fig:app-fine-tuned}
\end{figure}

\begin{figure}[ht]
    \centering
    \includegraphics[width=0.98\linewidth]{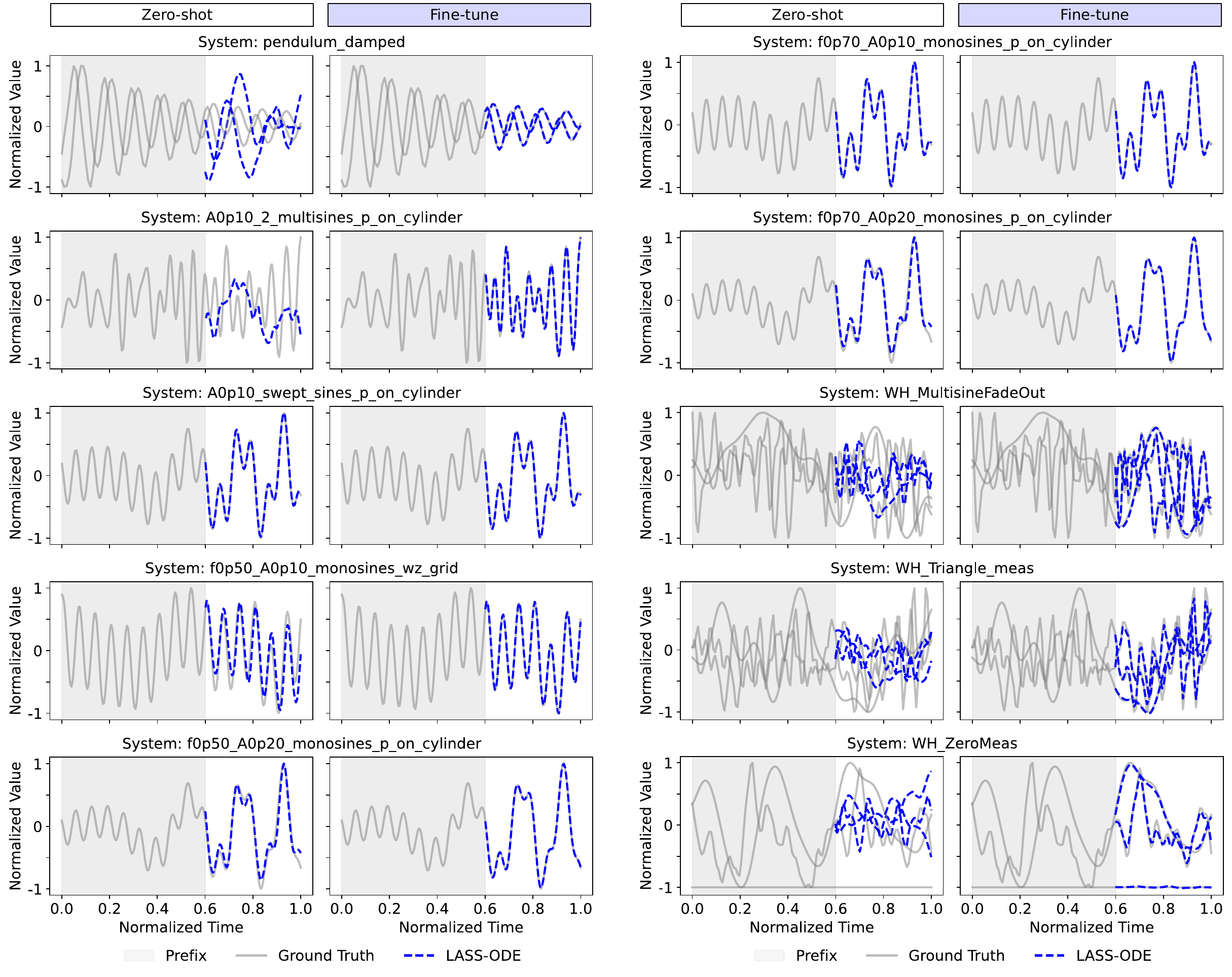}
    \caption{Performance comparison of the same base system under zero-shot inference (left) and after LoRA fine-tuning (right) for 30 epochs. The prefix ratio is set to $60\%$.}
    \label{fig:app-fine-tuned-2}
\end{figure}

In this section, we show the complete zero-shot tests on systems that never appear in the training dataset. Table \ref{tab:zero-shot-MSE-app} reports the results. For most of the systems, LASS-ODE can directly output accurate predictions. However, there are also systems that are hard to fit. All methods nearly fail in these systems. However, we find that with around $30$ epochs of fine-tuning, LASS-ODE can quickly be able to capture the underlying ODE structure and achieve accurate extrapolation, as shown in Figs. \ref{fig:app-fine-tuned} and \ref{fig:app-fine-tuned-2}.

\begin{table*}[ht]
\caption{
Test MSE ($\times 10^{-2}$) for zero-shot systems.
}

\label{tab:zero-shot-MSE-app}
\centering
\tiny
\begin{tabular}{p{1.4cm}*{27}{p{0.11cm}}}
\toprule
\multirow{2}{*}[0.4em]{Models} & 
\multicolumn{3}{c}{\makecell{\textbf{LASS-ODE}\\(Ours)}} &
\multicolumn{3}{c}{\makecell{\textbf{TimesFM}\\\citeyearpar{das2024decoder}}} & 
\multicolumn{3}{c}{\makecell{\textbf{Chronos}\\\citeyearpar{ansari2024chronos}}} & 
\multicolumn{3}{c}{\makecell{\textbf{Timer-XL}\\\citeyearpar{liu2024timer}}} &
\multicolumn{3}{c}{\makecell{\textbf{Informer}\\\citeyearpar{zhou2021informer}}} &
\multicolumn{3}{c}{\makecell{\textbf{Autoformer}\\\citeyearpar{wu2021autoformer}}} &
\multicolumn{3}{c}{\makecell{\textbf{Latent ODE}\\\citeyearpar{rubanova2019latent}}} &
\multicolumn{3}{c}{\makecell{\textbf{Latent MoS}\\\citeyearpar{li2025latent}}} &
\multicolumn{3}{c}{\makecell{\textbf{ContiFormer}\\\citeyearpar{chen2023contiformer}}} \\

\cmidrule(lr){2-4}\cmidrule(lr){5-7}\cmidrule(lr){8-10}\cmidrule(lr){11-13}\cmidrule(lr){14-16}\cmidrule(lr){17-19}\cmidrule(lr){20-22}\cmidrule(lr){23-25}\cmidrule(lr){26-28}

Prefix ratio&
30\% & 60\% & 90\% &
30\% & 60\% & 90\% &
30\% & 60\% & 90\% &
30\% & 60\% & 90\% &
30\% & 60\% & 90\% &
30\% & 60\% & 90\% &
30\% & 60\% & 90\% &
30\% & 60\% & 90\% &
30\% & 60\% & 90\% \\
\midrule
\input{zero-shot_app.tex}
\end{tabular}
\end{table*}

\end{document}

%% file: in_domain.tex
\multirow{3}{*}{Cauchy\_euler} & 30.0\% & {\bf 0.012} & 74.06 & 69.04 & 84.44 & 33.38 & 19.00 & 0.627 & 0.297 & 20.16 \\
 & 60.0\% & {\bf 0.005} & 13.80 & 27.66 & 23.93 & 57.22 & 32.58 & 0.543 & 0.144 & 16.85 \\
 & 90.0\% & {\bf 0.007} & 0.322 & 0.063 & 0.639 & 67.11 & 6.16 & 0.683 & 0.058 & 1.65 \\
 & Avg. & {\bf 0.008} & 29.39 & 32.25 & 36.33 & 52.57 & 19.25 & 0.618 & 0.166 & 12.89 \\
\midrule
\multirow{3}{*}{Fitzhugh\_nagumo} & 30.0\% & {\bf 0.017} & 97.83 & 136.04 & 136.72 & 48.78 & 79.06 & 0.909 & 0.519 & 33.15 \\
 & 60.0\% & {\bf 0.014} & 28.08 & 52.27 & 64.29 & 48.86 & 19.32 & 0.814 & 0.267 & 22.13 \\
 & 90.0\% & {\bf 0.013} & 3.09 & 1.38 & 11.98 & 53.29 & 4.86 & 0.832 & 0.127 & 5.38 \\
 & Avg. & {\bf 0.014} & 43.00 & 63.23 & 71.00 & 50.31 & 34.41 & 0.852 & 0.304 & 20.22 \\
\midrule
\multirow{3}{*}{Sweptsines\_pc1} & 30.0\% & {\bf 0.094} & 30.39 & 76.59 & 36.58 & 38.02 & 2.21 & 1.89 & 0.560 & 27.60 \\
 & 60.0\% & {\bf 0.091} & 30.58 & 5.01 & 38.20 & 37.56 & 2.75 & 1.85 & 0.575 & 9.19 \\
 & 90.0\% & {\bf 0.086} & 13.76 & 5.51 & 38.12 & 55.77 & 4.06 & 2.19 & 0.512 & 11.56 \\
 & Avg. & {\bf 0.090} & 24.91 & 29.04 & 37.63 & 43.78 & 3.01 & 1.97 & 0.549 & 16.12 \\
\midrule
\multirow{3}{*}{Wh\_sineinmeas} & 30.0\% & {\bf 0.077} & 69.17 & 56.02 & 95.99 & 31.49 & 81.79 & 1.55 & 1.21 & 21.72 \\
 & 60.0\% & {\bf 0.060} & 43.38 & 30.07 & 44.52 & 39.66 & 43.33 & 1.55 & 0.839 & 18.56 \\
 & 90.0\% & {\bf 0.065} & 9.75 & 6.18 & 13.47 & 58.99 & 20.81 & 1.95 & 0.559 & 12.10 \\
 & Avg. & {\bf 0.067} & 40.77 & 30.75 & 51.33 & 43.38 & 48.64 & 1.68 & 0.868 & 17.46 \\
\midrule
\multicolumn{11}{c}{Zero-shot Test}\\
\midrule
\input{zero-shot.tex}

%% file: zero-shot.tex
\multirow{3}{*}{Sweptsines\_pc4} & 30.0\% & \bf{0.043} & 19.81 & 51.98 & 22.31 & 20.82 & 16.64 & 0.944 & 0.481 & 16.17 \\
 & 60.0\% & \bf{0.037} & 30.60 & 23.06 & 27.51 & 29.83 & 23.54 & 1.05 & 0.518 & 13.79 \\
 & 90.0\% & \bf{0.038} & 23.49 & 34.57 & 25.88 & 49.09 & 30.16 & 1.36 & 0.519 & 20.82 \\
 & Avg. & \bf{0.039} & 24.64 & 36.54 & 25.23 & 33.25 & 23.45 & 1.12 & 0.506 & 16.93 \\
\midrule
\multirow{3}{*}{Monosines\_pc3} & 30.0\% & \bf{0.045} & 19.76 & 29.56 & 22.30 & 20.79 & 16.64 & 0.965 & 0.494 & 12.95 \\
 & 60.0\% & \bf{0.038} & 30.55 & 26.36 & 27.47 & 30.61 & 23.34 & 1.08 & 0.524 & 14.77 \\
 & 90.0\% & \bf{0.038} & 23.31 & 38.02 & 25.75 & 51.48 & 29.98 & 1.40 & 0.521 & 22.17 \\
 & Avg. & \bf{0.040} & 24.54 & 31.31 & 25.17 & 34.29 & 23.32 & 1.15 & 0.513 & 16.63 \\
\midrule
\multirow{3}{*}{Multisines\_pc4} & 30.0\% & 24.89 & \bf{16.61} & 50.03 & 23.52 & 24.83 & 35.71 & 24.86 & 24.68 & 32.87 \\
 & 60.0\% & 25.85 & \bf{16.51} & 26.83 & 20.24 & 24.97 & 21.80 & 25.41 & 22.84 & 25.79 \\
 & 90.0\% & 25.59 & \bf{14.99} & 28.62 & 35.71 & 54.35 & 38.65 & 37.30 & 24.97 & 39.00 \\
 & Avg. & 25.44 & \bf{16.04} & 35.16 & 26.49 & 34.72 & 32.06 & 29.19 & 24.16 & 32.55 \\
\midrule
\multirow{3}{*}{Duffing} & 30.0\% & 45.26 & 67.71 & 80.24 & \bf{32.67} & 20.46  & 40.22 & 38.45 & 47.91 & 48.35 \\
 & 60.0\% & 43.91 & 46.77 & 56.98 & \bf{31.43} & 39.72 & 43.61 & 37.15 & 44.41 & 41.23 \\
 & 90.0\% & 44.74 & \bf{13.98} & 26.46 & 32.54 & 60.41 & 19.17 & 38.15 & 29.92 & 30.93 \\
 & Avg. & 44.63 & 42.82 & 54.56 & \bf{32.22} & 40.20 & 34.33 & 37.92 & 40.75 & 40.17 \\
\bottomrule

%% file: fine-tune.tex
\multirow{3}{*}{Multisines\_pc4} 
 & 30.0\%       & {\bf 0.197} & 1.882 & 8.473  & 6.006  \\
 & 60.0\%       & {\bf 0.194} & 1.779 & 7.212  & 4.816  \\
 & 90.0\%       & {\bf 0.191} & 1.693 & 1.853  & 3.248  \\
 & Avg.         & {\bf 0.194} & 1.785 & 5.846  & 4.690  \\
\midrule
\multirow{3}{*}{Duffing}         
 & 30.0\%       & {\bf 0.218} & 6.415  & 11.20  & 11.75  \\
 & 60.0\%       & {\bf 0.182} & 6.783  & 7.234  & 10.01  \\
 & 90.0\%       & {\bf 0.194} & 3.134  & 3.477  & 5.910  \\
 & Avg.         & {\bf 0.198} & 5.444  & 7.304  & 9.223  \\
 \bottomrule

%% file: in_domain_app.tex
Cauchy\_euler & {\bf 0.01} & {\bf 0.01} & {\bf 0.01} & 74.1 & 13.8 & 0.32 & 69.0 & 27.7 & 0.06 & 84.4 & 23.9 & 0.64 & 33.4 & 57.2 & 67.1 & 19.0 & 32.6 & 6.2 & 0.63 & 0.54 & 0.68 & 0.30 & 0.14 & 0.06 & 20.2 & 16.9 & 1.7 \\
Droop\_governor & {\bf 0.01} & {\bf 0.01} & {\bf 0.01} & 4.2 & 1.6 & 3.0 & 33.8 & 17.4 & 0.69 & 45.3 & 8.5 & 1.6 & 3.0 & 5.1 & 5.0 & 1.8 & 3.2 & 0.96 & 0.16 & 0.16 & 0.16 & 0.09 & 0.06 & 0.05 & 4.4 & 4.2 & 1.1 \\
Ermakov\_pinney & {\bf 0.06} & {\bf 0.05} & {\bf 0.05} & 10.7 & 3.6 & 1.9 & 23.7 & 2.5 & 0.47 & 67.6 & 51.8 & 37.6 & 51.5 & 51.2 & 51.0 & 9.2 & 6.3 & 4.3 & 1.7 & 1.6 & 1.7 & 0.46 & 0.32 & 0.27 & 19.2 & 7.7 & 3.9 \\
Hopf\_normal & {\bf 0.03} & 0.03 & {\bf 0.03} & 1.7 & {\bf 0.02} & 0.04 & 6.7 & 0.66 & 0.08 & 57.2 & 48.5 & 38.6 & 49.9 & 50.1 & 49.7 & 21.1 & 8.2 & 1.3 & 1.2 & 1.2 & 1.2 & 0.24 & 0.08 & 0.07 & 10.5 & 4.2 & 1.8 \\
Lotka\_volterra & {\bf 0.07} & {\bf 0.03} & {\bf 0.04} & 51.2 & 35.3 & 24.4 & 297.8 & 43.0 & 1.0 & 54.9 & 54.5 & 21.5 & 57.6 & 58.7 & 59.4 & 43.3 & 49.2 & 20.1 & 1.9 & 1.4 & 1.5 & 0.91 & 0.57 & 0.48 & 56.4 & 24.5 & 5.6 \\
Vanderpol\_mu10 & {\bf 1.1} & {\bf 0.29} & {\bf 0.10} & 56.5 & 40.6 & 12.9 & 121.0 & 42.5 & 19.1 & 66.4 & 53.1 & 19.6 & 38.2 & 65.5 & 70.8 & 18.3 & 31.4 & 31.5 & 6.5 & 4.4 & 2.6 & 4.2 & 2.0 & 0.82 & 42.4 & 32.1 & 21.7 \\
Fitzhugh\_nagumo & {\bf 0.02} & {\bf 0.01} & {\bf 0.01} & 97.8 & 28.1 & 3.1 & 136.0 & 52.3 & 1.4 & 136.7 & 64.3 & 12.0 & 48.8 & 48.9 & 53.3 & 79.1 & 19.3 & 4.9 & 0.91 & 0.81 & 0.83 & 0.52 & 0.27 & 0.13 & 33.1 & 22.1 & 5.4 \\
Pendulum & {\bf 0.10} & {\bf 0.10} & {\bf 0.11} & 14.2 & 9.5 & 4.6 & 32.1 & 1.1 & 0.14 & 64.9 & 51.8 & 42.6 & 48.6 & 48.4 & 48.4 & 41.6 & 32.2 & 22.7 & 2.2 & 2.3 & 2.3 & 0.92 & 0.81 & 0.66 & 22.3 & 5.8 & 2.5 \\
Rossler\_3d & {\bf 0.07} & {\bf 0.05} & {\bf 0.05} & 20.2 & 17.0 & 6.5 & 84.3 & 35.4 & 16.5 & 34.5 & 34.0 & 19.0 & 49.8 & 49.7 & 50.0 & 11.6 & 8.5 & 5.5 & 1.9 & 1.6 & 1.6 & 0.61 & 0.45 & 0.34 & 31.9 & 21.8 & 16.1 \\
Smib\_swing & {\bf 0.02} & {\bf 0.01} & {\bf 0.02} & 41.0 & 7.1 & 0.24 & 68.6 & 35.7 & 0.69 & 52.5 & 10.2 & 0.96 & 23.2 & 8.9 & 7.2 & 50.8 & 25.7 & 0.42 & 0.61 & 0.36 & 0.34 & 0.39 & 0.22 & 0.05 & 17.3 & 8.2 & 1.5 \\
Vanderpol\_mu0.5 & {\bf 0.01} & {\bf 0.01} & 0.01 & 85.4 & 9.5 & 1.5 & 105.7 & 67.0 & {\bf 0.01} & 61.2 & 83.5 & 16.7 & 26.3 & 45.1 & 45.1 & 0.85 & 1.5 & 1.6 & 0.61 & 0.79 & 0.78 & 0.18 & 0.13 & 0.09 & 21.6 & 23.5 & 0.76 \\
Vanderpol\_mu10 & {\bf 1.1} & {\bf 0.29} & {\bf 0.10} & 56.5 & 40.6 & 12.9 & 121.0 & 42.5 & 19.1 & 66.4 & 53.1 & 19.6 & 38.2 & 65.5 & 70.8 & 18.3 & 31.4 & 31.5 & 6.5 & 4.4 & 2.6 & 4.2 & 2.0 & 0.82 & 42.4 & 32.1 & 21.7 \\
Vanderpol\_mu1 & {\bf 0.02} & {\bf 0.02} & {\bf 0.02} & 58.5 & 14.1 & 7.5 & 144.7 & 69.0 & 8.1 & 48.0 & 59.9 & 19.4 & 23.9 & 40.9 & 40.5 & 1.0 & 1.8 & 1.9 & 0.65 & 0.80 & 0.79 & 0.20 & 0.16 & 0.14 & 23.9 & 23.0 & 9.7 \\
Vanderpol\_mu3 & {\bf 0.03} & {\bf 0.03} & {\bf 0.03} & 58.2 & 44.3 & 8.4 & 65.6 & 63.8 & 7.8 & 50.8 & 35.6 & 11.6 & 20.6 & 35.5 & 35.0 & 3.9 & 6.6 & 2.0 & 0.77 & 0.95 & 0.96 & 0.35 & 0.34 & 0.20 & 17.0 & 21.7 & 9.4 \\
Lorenz\_3d & {\bf 0.08} & {\bf 0.06} & {\bf 0.07} & 39.3 & 33.5 & 19.6 & 80.2 & 62.5 & 12.5 & 35.9 & 28.0 & 19.2 & 27.2 & 29.1 & 27.6 & 26.1 & 26.8 & 17.7 & 1.5 & 1.3 & 1.4 & 0.87 & 0.72 & 0.64 & 23.4 & 21.3 & 11.0 \\
Sweptsines\_vx & {\bf 0.12} & {\bf 0.12} & {\bf 0.13} & 37.3 & 31.8 & 18.2 & 46.1 & 1.4 & 0.83 & 46.3 & 42.3 & 46.0 & 46.3 & 48.2 & 40.5 & 2.8 & 1.2 & 0.50 & 2.4 & 2.5 & 2.3 & 0.73 & 0.60 & 0.45 & 25.6 & 6.4 & 4.8 \\
Sweptsines\_vy & {\bf 0.12} & {\bf 0.12} & {\bf 0.13} & 35.4 & 31.3 & 16.1 & 44.7 & 0.95 & 0.47 & 51.6 & 42.2 & 41.1 & 45.8 & 48.1 & 52.3 & 1.3 & 1.7 & 0.73 & 2.4 & 2.4 & 2.6 & 0.61 & 0.62 & 0.47 & 25.1 & 5.5 & 4.3 \\
Sweptsines\_p & {\bf 0.12} & {\bf 0.12} & {\bf 0.12} & 37.3 & 32.6 & 15.6 & 64.8 & 5.8 & 2.2 & 41.8 & 39.7 & 35.3 & 43.9 & 43.7 & 53.7 & 1.4 & 1.6 & 0.77 & 2.3 & 2.3 & 2.5 & 0.62 & 0.62 & 0.45 & 28.5 & 10.8 & 8.1 \\
Sweptsines\_pc1 & {\bf 0.09} & {\bf 0.09} & {\bf 0.09} & 30.4 & 30.6 & 13.8 & 76.6 & 5.0 & 5.5 & 36.6 & 38.2 & 38.1 & 38.0 & 37.6 & 55.8 & 2.2 & 2.7 & 4.1 & 1.9 & 1.8 & 2.2 & 0.56 & 0.57 & 0.51 & 27.6 & 9.2 & 11.6 \\
Sweptsines\_wz & {\bf 0.09} & {\bf 0.09} & {\bf 0.10} & 24.7 & 29.1 & 11.2 & 39.0 & 41.7 & 2.2 & 31.6 & 26.5 & 31.3 & 32.7 & 34.4 & 25.0 & 7.6 & 6.9 & 6.0 & 1.7 & 1.7 & 1.6 & 0.68 & 0.66 & 0.57 & 19.5 & 20.4 & 5.5 \\
Multisines\_vx & {\bf 0.13} & {\bf 0.13} & {\bf 0.13} & 43.0 & 45.0 & 18.1 & 37.2 & 26.7 & 5.9 & 33.1 & 34.9 & 47.9 & 37.6 & 46.0 & 36.7 & 8.2 & 5.8 & 4.9 & 2.2 & 2.4 & 2.2 & 0.95 & 0.89 & 0.72 & 21.2 & 20.6 & 10.0 \\
Multisines\_vy & {\bf 0.12} & {\bf 0.13} & {\bf 0.13} & 55.1 & 38.2 & 12.1 & 52.0 & 25.5 & 14.2 & 34.1 & 32.8 & 38.8 & 32.1 & 34.9 & 53.2 & 6.6 & 9.1 & 6.0 & 2.0 & 2.1 & 2.6 & 0.92 & 0.95 & 0.68 & 22.3 & 17.6 & 17.2 \\
Multisines\_p & {\bf 0.09} & {\bf 0.09} & {\bf 0.09} & 36.0 & 29.3 & 5.9 & 37.3 & 35.6 & 8.5 & 26.7 & 23.9 & 17.0 & 29.5 & 35.3 & 25.8 & 9.8 & 23.6 & 2.6 & 1.7 & 1.8 & 1.5 & 0.78 & 0.87 & 0.41 & 18.2 & 19.5 & 9.4 \\
Multisines\_pc1 & {\bf 0.09} & {\bf 0.08} & {\bf 0.09} & 39.9 & 38.9 & 14.8 & 43.4 & 30.6 & 23.3 & 38.9 & 17.8 & 18.1 & 23.2 & 24.9 & 16.9 & 20.2 & 40.3 & 14.6 & 1.4 & 1.4 & 1.2 & 0.88 & 0.97 & 0.70 & 17.0 & 15.3 & 11.4 \\
Multisines\_wz & {\bf 0.14} & {\bf 0.14} & {\bf 0.13} & 28.6 & 39.6 & 19.0 & 42.2 & 29.8 & 13.1 & 24.1 & 18.7 & 24.2 & 25.6 & 27.9 & 18.7 & 19.7 & 16.5 & 3.0 & 1.9 & 2.0 & 1.6 & 1.1 & 1.1 & 0.68 & 18.6 & 16.8 & 9.9 \\
Multisines\_pc2 & {\bf 0.26} & {\bf 0.28} & {\bf 0.25} & 22.5 & 17.5 & 16.0 & 37.4 & 23.7 & 17.7 & 16.6 & 12.8 & 24.0 & 19.6 & 16.3 & 27.7 & 16.2 & 16.4 & 14.4 & 2.3 & 2.1 & 2.7 & 1.5 & 1.4 & 1.3 & 16.5 & 12.6 & 14.5 \\
Sweptsines\_pc2 & {\bf 0.07} & {\bf 0.07} & {\bf 0.06} & 24.9 & 27.7 & 15.8 & 111.3 & 10.2 & 9.8 & 29.4 & 31.7 & 30.5 & 29.9 & 30.8 & 49.9 & 5.6 & 7.8 & 10.7 & 1.4 & 1.4 & 1.8 & 0.54 & 0.58 & 0.54 & 27.6 & 10.7 & 13.4 \\
Sweptsines\_pc3 & {\bf 0.06} & {\bf 0.06} & {\bf 0.06} & 22.2 & 27.5 & 17.5 & 71.6 & 18.4 & 14.5 & 25.9 & 29.0 & 27.2 & 25.7 & 29.5 & 48.9 & 9.1 & 12.8 & 17.7 & 1.3 & 1.3 & 1.7 & 0.55 & 0.59 & 0.56 & 21.3 & 13.1 & 15.3 \\
Multisines\_pc3 & {\bf 0.26} & {\bf 0.28} & {\bf 0.25} & 22.5 & 17.5 & 16.0 & 37.3 & 26.0 & 16.6 & 16.6 & 12.8 & 24.0 & 19.3 & 16.2 & 28.5 & 16.2 & 16.4 & 14.4 & 2.3 & 2.1 & 2.7 & 1.5 & 1.4 & 1.3 & 16.4 & 13.0 & 14.3 \\
Monosines\_vx & {\bf 0.09} & {\bf 0.09} & {\bf 0.09} & 31.9 & 26.6 & 17.1 & 45.2 & 15.7 & 5.3 & 32.5 & 30.4 & 32.0 & 34.4 & 35.4 & 25.1 & 10.0 & 3.5 & 2.3 & 1.7 & 1.7 & 1.5 & 0.74 & 0.56 & 0.49 & 21.1 & 14.0 & 7.7 \\
Monosines\_vy & {\bf 0.12} & {\bf 0.12} & {\bf 0.12} & 28.4 & 25.2 & 15.5 & 28.9 & 4.1 & 2.6 & 41.6 & 34.5 & 33.5 & 39.1 & 40.1 & 52.7 & 2.5 & 2.8 & 2.9 & 2.2 & 2.2 & 2.5 & 0.66 & 0.66 & 0.59 & 19.4 & 9.0 & 8.6 \\
Monosines\_p & {\bf 0.07} & {\bf 0.07} & {\bf 0.07} & 25.6 & 33.5 & 14.9 & 243.7 & 18.4 & 13.4 & 25.3 & 26.2 & 19.0 & 36.4 & 34.1 & 44.1 & 12.0 & 8.1 & 3.9 & 1.6 & 1.6 & 1.8 & 0.66 & 0.63 & 0.46 & 41.9 & 14.4 & 14.4 \\
Monosines\_pc1 & {\bf 0.06} & {\bf 0.05} & {\bf 0.05} & 20.9 & 28.3 & 19.1 & 56.4 & 25.8 & 21.6 & 24.1 & 28.0 & 25.6 & 23.5 & 28.4 & 48.8 & 12.2 & 17.0 & 22.7 & 1.2 & 1.2 & 1.5 & 0.54 & 0.58 & 0.55 & 18.3 & 14.5 & 17.7 \\
Wh\_combined0 & {\bf 0.22} & {\bf 0.21} & {\bf 0.22} & 42.1 & 30.4 & 16.7 & 57.6 & 38.5 & 19.3 & 36.6 & 24.2 & 17.5 & 24.9 & 25.1 & 25.7 & 35.9 & 26.9 & 24.7 & 2.3 & 2.3 & 2.4 & 1.7 & 1.5 & 1.3 & 21.7 & 18.4 & 14.2 \\
Wh\_estimation & {\bf 0.23} & {\bf 0.22} & {\bf 0.23} & 34.3 & 34.7 & 20.1 & 58.3 & 42.2 & 20.4 & 30.0 & 26.8 & 19.4 & 24.1 & 24.3 & 25.6 & 26.3 & 29.1 & 25.4 & 2.3 & 2.3 & 2.4 & 1.6 & 1.6 & 1.4 & 21.5 & 18.9 & 14.6 \\
Wh\_sineinmeas & {\bf 0.08} & {\bf 0.06} & {\bf 0.06} & 69.2 & 43.4 & 9.7 & 56.0 & 30.1 & 6.2 & 96.0 & 44.5 & 13.5 & 31.5 & 39.7 & 59.0 & 81.8 & 43.3 & 20.8 & 1.6 & 1.5 & 2.0 & 1.2 & 0.84 & 0.56 & 21.7 & 18.6 & 12.1 \\
Wh\_sinesweepin & {\bf 0.05} & {\bf 0.03} & {\bf 0.04} & 62.5 & 30.3 & 3.1 & 134.0 & 38.6 & 3.4 & 136.8 & 43.7 & 12.0 & 62.5 & 64.9 & 65.0 & 50.6 & 36.4 & 33.9 & 1.8 & 1.5 & 1.6 & 0.84 & 0.52 & 0.36 & 41.6 & 24.7 & 9.5 \\
Wh\_triangle2 & {\bf 0.29} & {\bf 0.28} & {\bf 0.28} & 32.0 & 27.7 & 17.1 & 43.5 & 32.5 & 20.4 & 32.4 & 22.4 & 16.5 & 23.4 & 23.0 & 20.8 & 26.3 & 27.9 & 23.9 & 2.6 & 2.5 & 2.4 & 1.8 & 1.8 & 1.6 & 19.3 & 17.0 & 13.4 \\
\bottomrule

%% file: zero-shot_app.tex
Duffing & 45.3 & 43.9 & 44.7 & 67.7 & 46.8 & {\bf 14.0} & 80.2 & 57.0 & 26.5 & {\bf 32.7} & {\bf 31.4} & 32.5 & 20.5 & 39.7 & 60.4 & 40.2 & 43.6 & 19.2 & 38.5 & 37.2 & 38.2 & 47.9 & 44.4 & 29.9 & 48.4 & 41.2 & 30.9 \\
Emden\_fowler & {\bf 7.3} & {\bf 3.8} & {\bf 2.4} & 34.3 & 4.3 & 3.6 & 69.5 & 16.5 & 9.4 & 60.1 & 12.8 & 0.89 & 33.5 & 57.5 & 58.0 & 9.9 & 16.9 & 3.6 & 15.6 & 19.7 & 19.3 & 10.6 & 7.4 & 3.2 & 38.5 & 28.2 & 3.5 \\
Hamiltonian\_sho & {\bf 0.97} & {\bf 0.91} & {\bf 0.86} & 29.3 & 30.2 & 10.6 & 58.8 & 39.1 & 20.2 & 74.2 & 54.2 & 23.0 & 50.6 & 50.5 & 50.5 & 6.1 & 4.2 & 2.8 & 70.2 & 70.0 & 70.0 & 17.5 & 10.2 & 8.2 & 57.4 & 10.8 & 2.4 \\
Pendulum\_damped & 21.5 & 20.8 & 20.5 & 19.5 & 7.8 & 2.1 & 48.7 & 11.7 & {\bf 0.96} & 19.3 & 8.0 & 1.6 & 8.2 & 4.4 & 3.1 & {\bf 5.9} & {\bf 2.2} & 2.9 & 13.3 & 9.6 & 7.9 & 16.3 & 10.9 & 8.8 & 18.4 & 7.6 & 2.3 \\
Multisines\_pc4 & 24.9 & 25.9 & 25.6 & {\bf 16.6} & {\bf 16.5} & {\bf 15.0} & 50.0 & 26.8 & 28.6 & 23.5 & 20.2 & 35.7 & 24.8 & 25.0 & 54.4 & 35.7 & 21.8 & 38.7 & 24.9 & 25.4 & 37.3 & 24.7 & 22.8 & 25.0 & 32.9 & 25.8 & 39.0 \\
Sweptsines\_pc4 & {\bf 0.04} & {\bf 0.04} & {\bf 0.04} & 19.8 & 30.6 & 23.5 & 52.0 & 23.1 & 34.6 & 22.3 & 27.5 & 25.9 & 20.8 & 29.8 & 49.1 & 16.6 & 23.5 & 30.2 & 0.94 & 1.1 & 1.4 & 0.48 & 0.52 & 0.52 & 16.2 & 13.8 & 20.8 \\
Monosines\_wz & {\bf 2.3} & {\bf 2.4} & {\bf 2.3} & 21.6 & 27.4 & 16.0 & 52.6 & 30.7 & 12.6 & 27.1 & 23.9 & 14.8 & 28.2 & 25.7 & 31.5 & 14.1 & 13.1 & 3.1 & 8.0 & 7.8 & 8.5 & 5.1 & 5.5 & 3.6 & 28.1 & 21.8 & 16.8 \\
Monosines\_pc2 & {\bf 0.42} & {\bf 0.44} & {\bf 0.45} & 22.5 & 38.5 & 30.8 & 34.0 & 42.2 & 32.2 & 24.0 & 31.4 & 27.7 & 21.8 & 31.6 & 60.1 & 24.4 & 36.8 & 33.6 & 3.0 & 3.7 & 5.2 & 2.1 & 2.6 & 2.5 & 17.6 & 23.1 & 28.7 \\
Monosines\_pc3 & {\bf 0.04} & {\bf 0.04} & {\bf 0.04} & 19.8 & 30.6 & 23.3 & 29.6 & 26.4 & 38.0 & 22.3 & 27.5 & 25.8 & 20.8 & 30.6 & 51.5 & 16.6 & 23.3 & 30.0 & 0.96 & 1.1 & 1.4 & 0.49 & 0.52 & 0.52 & 13.0 & 14.8 & 22.2 \\
Monosines\_pc4 & {\bf 0.42} & {\bf 0.44} & {\bf 0.45} & 22.5 & 38.5 & 30.8 & 28.4 & 36.2 & 38.0 & 24.0 & 31.4 & 27.7 & 21.8 & 34.4 & 60.1 & 24.4 & 36.9 & 33.6 & 3.0 & 3.9 & 5.2 & 2.1 & 2.6 & 2.5 & 16.3 & 22.7 & 30.7 \\
Wh\_multisine & {\bf 20.7} & {\bf 21.2} & {\bf 22.3} & 46.5 & 39.2 & 26.2 & 42.5 & 40.7 & 23.6 & 49.0 & 28.6 & 28.8 & 23.5 & 24.6 & 24.0 & 44.0 & 23.5 & 36.2 & 25.1 & 25.4 & 24.8 & 33.0 & 27.8 & 26.5 & 30.2 & 30.3 & 24.0 \\
Wh\_triangle & 26.0 & 26.0 & 25.8 & 31.6 & 27.9 & {\bf 19.5} & 33.4 & 30.1 & 31.9 & 32.7 & 24.0 & 21.9 & {\bf 22.8} & {\bf 22.6} & 26.2 & 28.3 & 33.5 & 27.4 & 24.3 & 24.3 & 26.0 & 27.5 & 27.8 & 24.7 & 26.9 & 25.7 & 28.3 \\
Wh\_zeromeas & 15.4 & {\bf 14.5} & {\bf 15.9} & 27.8 & 35.1 & {\bf 8.7} & 36.9 & 35.6 & 16.7 & 39.5 & 34.1 & 31.0 & 51.8 & 51.7 & 46.4 & 22.1 & 31.1 & 14.2 & 46.8 & 47.9 & 46.2 & 34.2 & 39.5 & 26.0 & 44.3 & 43.9 & 30.8\\
\bottomrule